\documentclass[10pt,twocolumn,letterpaper]{article}
\usepackage[pagenumbers]{iccv} 
\usepackage{iccv}
\usepackage{times}
\usepackage{epsfig}
\usepackage{graphicx}
\usepackage{amsmath}
\usepackage{amssymb}

\usepackage{amsmath,amsfonts}
\usepackage{algorithmic}
\usepackage{algorithm}
\usepackage{array}
\usepackage{textcomp}
\usepackage{stfloats}
\usepackage{url}
\usepackage{verbatim}
\usepackage{graphicx}
\usepackage{graphicx}
\usepackage{amsmath}
\usepackage{amssymb}
\usepackage{booktabs}
\usepackage{multirow}
\usepackage{algorithm}
\usepackage{algorithmic}

\usepackage{setspace}

\usepackage{colortbl}
\usepackage{xcolor}
\usepackage{array} 
\usepackage{float}
\usepackage{arydshln}

\usepackage{svg}
\usepackage{graphicx}
\usepackage{caption,setspace}

\definecolor{iccvblue}{rgb}{0.21,0.49,0.74}
\usepackage[pagebackref,breaklinks,colorlinks,allcolors=iccvblue]{hyperref}

\begin{document}

\title{
The Devil is in the Darkness: Diffusion-Based Nighttime Dehazing Anchored in Brightness Perception 
}

\author{Xiaofeng Cong\textsuperscript{*} \\
Southeast University \\
{\tt\small cxf\_svip@163.com}
\and
Yu-Xin Zhang\textsuperscript{*} \\
Southeast University\\
{\tt\small yuxinzhang@seu.edu.cn}
\and
Haoran Wei \\
UESTC \\
{\tt\small hrwei@std.uestc.edu.cn}
\and
Yeying Jin \\
Tencent Company \\
{\tt\small e0178303@u.nus.edu}
\and
Junming Hou \\
Southeast University\\
{\tt\small junming\_hou@seu.edu.cn}
\and
Jie Gui\textsuperscript{\ddag} \\
Southeast University\\
{\tt\small guijie@seu.edu.cn}
\and
Jing Zhang\textsuperscript{\ddag} \\
Wuhan University\\
{\tt\small jingzhang.cv@gmail.com}
\and
Dacheng Tao \\
Nanyang Technological University\\
{\tt\small dacheng.tao@gmail.com}
}

\maketitle

\renewcommand{\thefootnote}{} 
\footnotetext{* equal contributions, \ddag \hspace{0.05cm} corresponding authors}
\renewcommand{\thefootnote}{\arabic{footnote}}

\begin{abstract}
While nighttime image dehazing has been extensively studied, converting nighttime hazy images to daytime-equivalent brightness remains largely unaddressed. Existing methods face two critical limitations: (1) datasets overlook the brightness relationship between day and night, resulting in the brightness mapping being inconsistent with the real world during image synthesis; and (2) models do not explicitly incorporate daytime brightness knowledge, limiting their ability to reconstruct realistic lighting. To address these challenges, we introduce the Diffusion-Based Nighttime Dehazing (DiffND) framework, which excels in both data synthesis and lighting reconstruction. Our approach starts with a data synthesis pipeline that simulates severe distortions while enforcing brightness consistency between synthetic and real-world scenes, providing a strong foundation for learning night-to-day brightness mapping. Next, we propose a restoration model that integrates a pre-trained diffusion model guided by a brightness perception network. This design harnesses the diffusion model's generative ability while adapting it to nighttime dehazing through brightness-aware optimization. Experiments validate our dataset's utility and the model's superior performance in joint haze removal and brightness mapping. 
\end{abstract}

\section{Introduction}
\label{sec:introduction}
Due to the absorption and scattering of light, haze effect may be pronounced in nighttime scenes with artificial light sources, which will degrade the quality of the image \cite{liu2021single,kuanar2022multi,zhang2014nighttime,banerjee2021nighttime,jin2025raindrop}. Early attention was focused on the nighttime dehazing task (ND-Task) \cite{jin2023enhancing}, in which the brightness of the dehazed image remains unchanged overall \cite{zhang2020nighttime,cong2024semi,lin2024nighthaze,chen2024ni}.

In nighttime scenes, even if the haze is removed, the visibility of the scene is still limited, since the dehazed image is still in darkness \cite{zhang2020nighttime}. The semantics of the scene may be difficult to understand under a dark environment \cite{kennerley20232pcnet}. Meanwhile, the color distribution of the dehazed images under darkness is still affected by ambient lights \cite{jin2023enhancing}. Consequently, low-light image enhancement algorithms cannot be directly applied as post-processing techniques to simply elevate the dehazed image to a daytime brightness level \cite{liu2022single}. 

Therefore, a more challenging topic is explored by Liu et al. \cite{liu2023nighthazeformer}, that is, \textit{the nighttime dehazing and brightness mapping task (NDBM-Task), whose purpose is to simultaneously achieve haze removal and brightness mapping from nighttime to daytime.} Under this purpose, the visibility of the image will be significantly improved. From the perspective of optimization objectives, the difference between the two tasks lies in the labels used during training. Algorithms that use \textit{nighttime darkness labels} belong to the ND-Task \cite{jin2023enhancing,liao2018hdp,cong2024semi}, while methods that adopt \textit{daytime brightness labels} belong to the NDBM-Task \cite{liu2023nighthazeformer,cui2023focal,cui2024omni}.

\begin{figure}
   \scriptsize
    \centering

   \includegraphics[width=2cm,height=1.3cm]{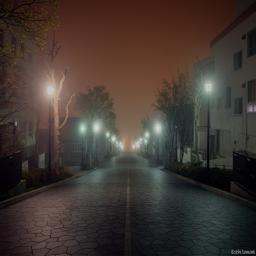}
   \includegraphics[width=2cm,height=1.3cm]{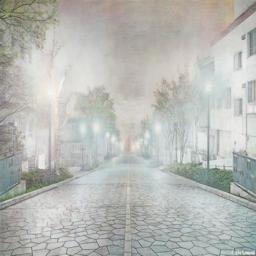}
   \includegraphics[width=2cm,height=1.3cm]{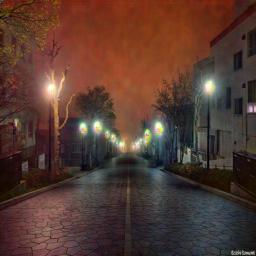}
   \includegraphics[width=2cm,height=1.3cm]{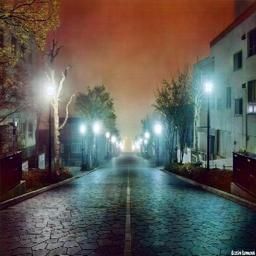}

   \includegraphics[width=2cm,height=1.3cm]{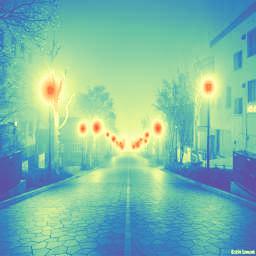}
   \includegraphics[width=2cm,height=1.3cm]{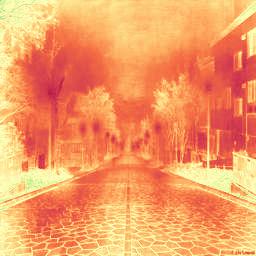}
   \includegraphics[width=2cm,height=1.3cm]{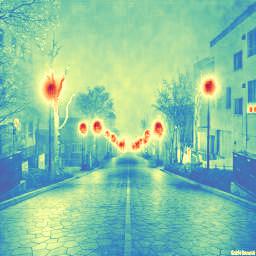}
   \includegraphics[width=2cm,height=1.3cm]{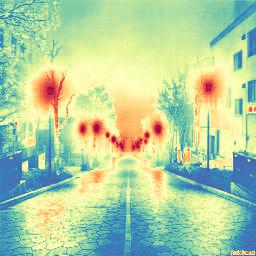}
  
   \leftline{\hspace{0.7cm} Hazy \hspace{0.8cm}  UNREAL-NH \cite{liu2023nighthazeformer} \hspace{0.05cm} NHR-Daytime \cite{zhang2020nighttime} \hspace{0.0cm}  NHM-Daytime \cite{zhang2020nighttime}}
   \vspace{0.1cm}

   \includegraphics[width=2cm,height=1.3cm]{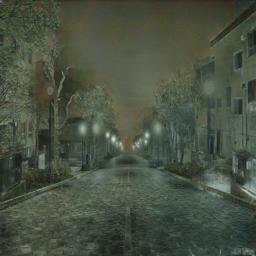}
   \includegraphics[width=2cm,height=1.3cm]{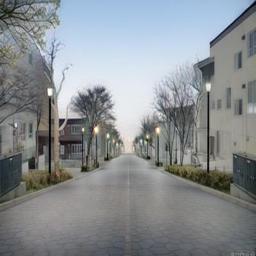}
   \includegraphics[width=0.1cm,height=1.3cm]{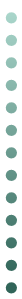}
   \includegraphics[width=1.9cm,height=1.3cm]{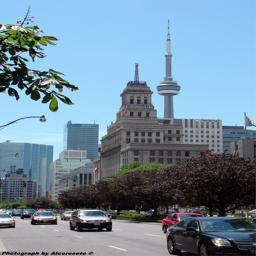}
   \includegraphics[width=1.9cm,height=1.3cm]{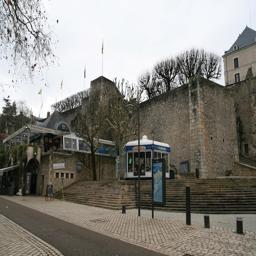}

   \includegraphics[width=2cm,height=1.3cm]{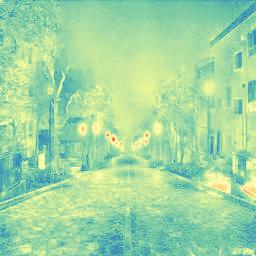}
   \includegraphics[width=2cm,height=1.3cm]{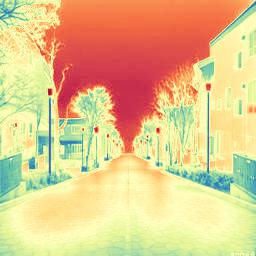}
   \includegraphics[width=0.1cm,height=1.3cm]{00_figures/plot_vertical_line/plot_vertical_line.png}
   \includegraphics[width=1.9cm,height=1.3cm]{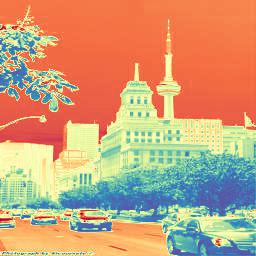}
   \includegraphics[width=1.9cm,height=1.3cm]{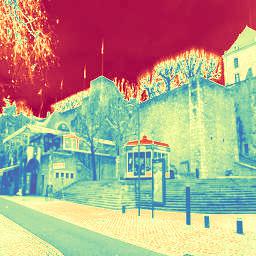}

   \leftline{\hspace{0.1cm}  NHC-Daytime \cite{zhang2020nighttime} \hspace{0.7cm}  Ours \hspace{1.8cm} Real-world Daytime}

    \vspace{-2mm}
   \caption{Performance on the real-world nighttime dehazing and brightness mapping task. The \textbf{daytime brightness labels} in UNREAL-NH and NH-series are used for the training process, which follows the settings in \cite{cui2024omni,cong2024semi}. The pseudo-color images denote the mean value of each channel. The results show that only our DiffND framework can achieve night-to-day brightness mapping with satisfactory lighting reconstruction. }
    \label{fig:motivation_of_this_paper}
\end{figure}

\begin{figure}
   \centering
   \footnotesize

  \includegraphics[width=\linewidth]{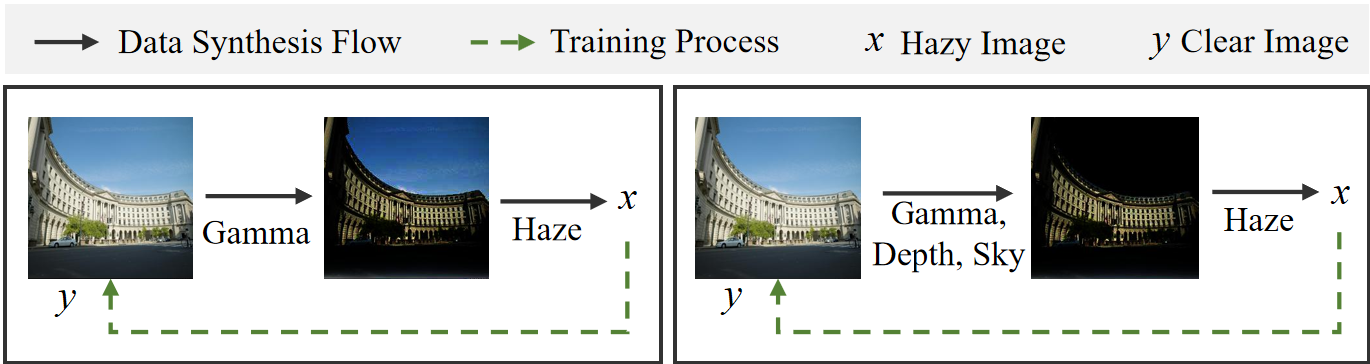}

   \leftline{\hspace{0.7cm} (a) existing datasets \cite{zhang2020nighttime}  \hspace{1.8cm} (b) our dataset}
   
   \caption{Comparison of data synthesis procedures. The daytime datasets in the NH-series do not utilize the depth and sky information during the Gamma correction \cite{zhang2020nighttime}, which leads to a globally uniform night-to-day brightness mapping. Our dataset builds a non-uniform night-to-day brightness mapping under the guidance of depth and sky, which is more consistent with the real world.}
   \label{fig:comparsion_of_brightness_in_differ_datasets}
 \end{figure}

Evaluations on synthetic data using daytime brightness labels show that well-designed networks can achieve high quantitative metrics in NDBM-Task \cite{cui2023focal,cui2024omni}. However, as shown  in Fig.~\ref{fig:motivation_of_this_paper}, the obtained results in the real-world evaluations are not visually satisfactory, which is due to:
\begin{itemize}
   \item \textbf{The datasets have inconsistent brightness mapping with the real world.} Due to the domain discrepancy caused by the game engine, the brightness mapping in UNREAL-NH differs from the real world \cite{liu2023nighthazeformer,cong2024semi}. Meanwhile, the synthesis process of NH-series \cite{zhang2020nighttime} adopts a global uniform brightness adjustment strategy as shown in Fig.~\ref{fig:comparsion_of_brightness_in_differ_datasets}. This makes the brightness mapping of the NH-series inconsistent with real-world non-uniformity nature, such as differences between sky and non-sky regions.
   \item \textbf{The models do not incorporate any prior knowledge about daytime brightness.} The models \cite{cong2024semi,jin2023enhancing,liu2023nighthazeformer} are designed under the content characteristics or statistical priors of night lighting. They are not aware of the utilization of the daytime brightness properties, which may make them unsuitable for the NDBM-Task.
\end{itemize}

This inspires us to design 1) a data synthesis scheme that better matches real-world night-to-day brightness mapping, and 2) a restoration diffusion model guided by a brightness perception network for lighting reconstruction. To verify that both our data and model are necessary, we conduct cross-validations in Section~\ref{subsec:ablation_studies_of_our_dataset_and_model}. We perform 1) train our model on existing datasets and 2) train existing models using data generated by our data synthesis pipeline. Both settings failed to achieve visual-friendly performance. This suggests the necessity of further exploration from both the data and model perspectives.

Due to the non-uniformity of the brightness mapping process, a globally consistent brightness adjustment, as existing solutions do \cite{zhang2020nighttime} for the daytime image, should not be performed. Rather, we use depth estimation and sky segmentation as the guidance to achieve non-uniform brightness adjustments. Meanwhile, to simulate haze degradation, we construct a distortion simulation method based on the severe degradation model \cite{lin2024nighthaze}, which has been proven to be effective for handling haze. However, \cite{lin2024nighthaze} requires real-world distortions for self-supervised training, which makes it impractical to apply to our task. Therefore, we design an alternative algorithm that replaces the real-world distortion with simulated distortion. \textit{By using depth, sky masks, and the severe degradation model with simulated distortion, we develop a data synthesis pipeline as the foundation for learning night-to-day brightness mapping.}

Based on the above synthetic data, we design a restoration network for lighting reconstruction. On the one hand, we utilize a pre-trained generative diffusion model to incorporate prior knowledge of real-world daytime brightness \cite{parmar2024one,sauer2024adversarial,chung2024style} while leveraging its exceptional image generation capabilities. On the other hand, considering the non-uniform brightness of night scenes, we introduce a brightness perception network that can identify non-uniform brightness in nighttime hazy images. \textit{By jointly training the diffusion model with the brightness perception network, we obtain a dehazing model capable of reconstructing daytime lighting.}

Overall, the main contributions of this paper include:
\begin{itemize}
   \item We propose the DiffND framework, which includes a data synthesis pipeline for night-to-day brightness mapping and a diffusion-based nighttime dehazing model with brightness perception for lighting reconstruction. Real-world evaluations demonstrate that our method produces visually appealing results for the NDBM-Task.
   \item We present a night-to-day data synthesis pipeline that integrates simulated severe distortions for realizing non-uniform brightness mapping. By combining depth and sky information, our pipeline ensures the brightness mapping aligns with real-world scenarios, encoding the transition from a hazy night to a clear day in the data.
   \item We present a diffusion-based nighttime dehazing model guided by a brightness perception network to detect non-uniform brightness in nighttime hazy images. By incorporating the learned brightness prior into the decoding process of a pre-trained diffusion network, our model delivers high-quality daytime lighting reconstructions.
\end{itemize}

 \begin{figure*}
   \footnotesize
   \centering
   \includegraphics[width=\linewidth]{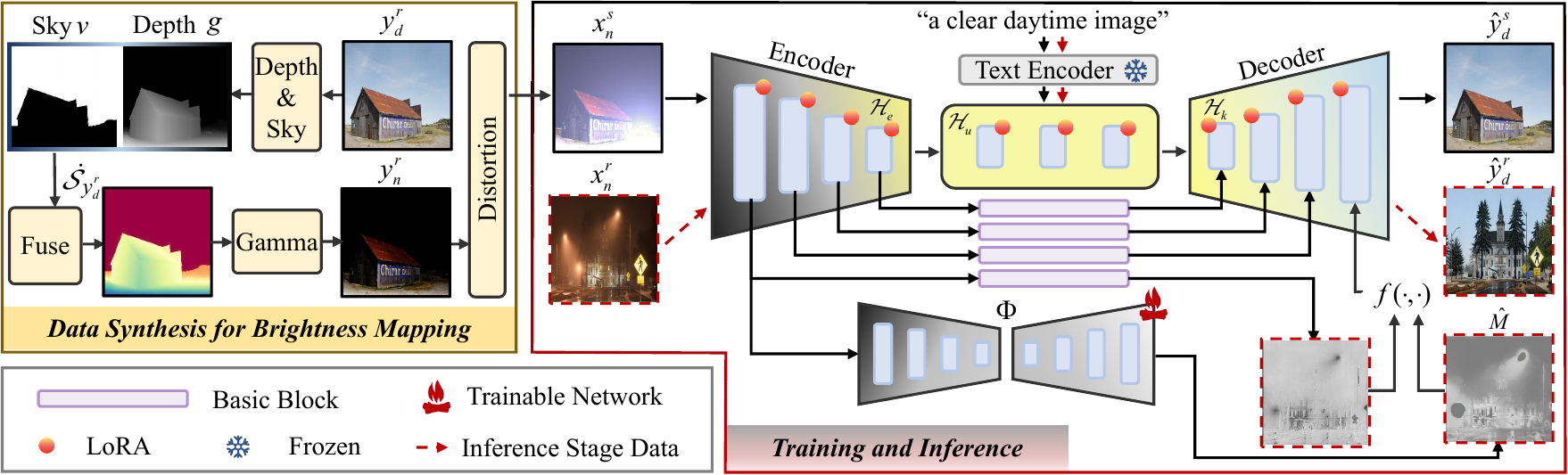}
   \caption{The data synthesis, network training, and inference processes. ``Distortion'' represents the simulated distortion.}
   \label{fig:data_synthesis_network_training_and_inference_process}
 \end{figure*}

\section{Related Work}
The research on nighttime haze includes two aspects \cite{yang2022variation,yu2019nighttime,li2015nighttime,wang2022variational,yang2018superpixel,liu2022multi,liu2022nighttime,lin2024nightrain,dai2022flare7k,jin2022unsupervised,jin2022structure}. The first is the ND-Task \cite{zhang2017fast}, which keeps the illumination of the image approximately unchanged. The second is NDBM-Task \cite{liu2023nighthazeformer}, which maps the brightness of the image to the daytime level.

\subsection{Dehazing Datasets}
Nighttime haze datasets include: (1) NH-series (NHR, NHM, NHC) \cite{zhang2020nighttime} with both nighttime and daytime labels, (2) UNREAL-NH \cite{liu2023nighthazeformer} with daytime labels, (3) GTA5 \cite{yan2020nighttime} with nighttime labels, (4) NightHaze $\&$ YellowHaze \cite{liao2018hdp} with nighttime labels. Depending on the labels, these datasets can be used for ND-Task and NDBM-Task, respectively. In this paper, our experimental results show that the datasets used for NDBM-Task can not effectively achieve brightness mapping from nighttime to daytime.

\subsection{Dehazing Methods}
From a data-driven perspective, the difference between ND-Task and NDBM-Task lies in the label images used during training processes \cite{liu2023nighthazeformer}. These two tasks use label images with nighttime darkness and daytime brightness images as optimization targets, respectively \cite{cong2024semi}. For neural networks \cite{zhou2023fourmer,cong2024semi} that can be trained in a supervised manner, they may be applied directly or indirectly to two tasks. 

The networks for handling the nighttime haze are mainly based on the convolutional, attention \cite{liu2023nighthazeformer}, Transformer \cite{liu2023nighthazeformer}, and Fourier operation \cite{cong2024semi}. The naive convolution is used to build HDPNet \cite{liao2018hdp}. NDNet \cite{zhang2020nighttime} adopts multi-scale encoder-decoder architectures. NightHazeFormer \cite{liu2023nighthazeformer} utilizes a multi-head self-attention mechanism to inject global and local information. Edge-aware convolution is designed by GAC \cite{jin2023enhancing}. Lin et al. \cite{lin2024nighthaze} propose a self-supervised dehazing convolutional framework based on two-stage training and fine-tuning. SFSNiD \cite{cong2024semi} builds spatial-frequency domain processing blocks with a multi-scale training strategy. Meanwhile, research has shown that supervised models used for daytime image dehazing tasks can also achieve excellent performance on nighttime haze datasets \cite{cong2024semi}. However, existing dehazing models lack the exploration of the knowledge of daytime brightness, which may limits the performance of lighting reconstructions.

\subsection{Generative Diffusion Models} 
The diffusion models \cite{yin2024one,ke2024repurposing} trained on a large amount of data demonstrate high-quality image priors, which make up for the difficulties faced by downstream tasks with insufficient data \cite{parmar2024one,yan2024diffusion,parmar2023zero,rombach2022high,xu2024prompt,si2024freeu}. \cite{sauer2024adversarial} proposes a pre-trained text-conditional one-step diffusion model SD-Turbo, which uses score distillation to leverage large-scale off-the-shelf image diffusion models as a teacher. CycleGAN-Turbo \cite{parmar2024one} uses the pre-trained SD-Turbo \cite{sauer2024adversarial} for image-to-image translations. In this paper, we leverage the prior knowledge of the pre-trained diffusion model \cite{parmar2024one,sauer2024adversarial} on daytime scenes to boost the lighting reconstruction performance.

\section{Methods}
Our DiffND consists of two main parts, namely, 1) a data synthesis pipeline for brightness mapping, and 2) a lighting reconstruction diffusion model incorporated with a brightness perception network. The data synthesis, network training and inference processes are shown in Fig.~\ref{fig:data_synthesis_network_training_and_inference_process}. $x$, $y$, and $\hat{y}$ represent the hazy, clear and predict dehazed image.  Subscripts $d$ and $n$ mean the illumination levels of daytime and nighttime. $s$ and $r$ indicate synthetic and real domains. For example, $x_{n}^{s}$ represents the synthetic ($s$) nighttime illumination level ($n$) of the hazy image ($x$). $p$ is the prompt \cite{li2024learning,tian2024argue}. $g$ and $v$ represent the depth information and sky mask respectively. $t$ and $\epsilon$ denote the time step and noise under the Gaussian distribution $\mathcal{N}(0, 1)$, respectively. The encoder, UNet \cite{sauer2024adversarial}, and decoder of pre-trained diffusion model \cite{parmar2024one} are denoted by $\mathcal{H}_{e}$,  $\mathcal{H}_{u}$, and $\mathcal{H}_{k}$.

\subsection{Data Synthesis for Brightness Mapping}

We use real-world daytime clear images to synthesize hazy nighttime images. A schematic diagram of the synthesis process is shown in Fig.~\ref{fig:data_synthesis_network_training_and_inference_process}. 

\noindent \textbf{Preliminary.} \cite{lin2024nighthaze} points out that distortions of night scenes are diverse, and an ideal physical imaging model may not be suitable for real-world scenes. The evaluations show that synthesizing nighttime haze using the linear severe degradation model \cite{lin2024nighthaze} can achieve effective dehazing performance on the ND-Task, which is:
\begin{equation}
   \label{eq:severe_distortion_model}
   x_{n}^{s} = W \odot y_{n}^{r} + (1 - W) \odot \mathcal{T} + \pi,
\end{equation}
where $\mathcal{T}$, $W$, and $\pi$ are the light, region weight, and noise. The $\odot$ is element-wise multiplication. However, Eq.~\ref{eq:severe_distortion_model} cannot be directly applied to our task. On the one hand, Eq.~\ref{eq:severe_distortion_model} does not need to achieve brightness mapping. On the other hand, there are too few severe distortion patches in the real world for our model training. Therefore, we replace $W \odot y_{n}^{r}$ with Eq.~\ref{eq:the_simulated_nighttime_clear_image} and $(1 - W) \odot \mathcal{T}$ with Eq. \ref{eq:light_centers}. By such modifications, the linear severe degradation model can be used in our task.

\noindent \textbf{Non-uniform brightness mapping.} In nighttime hazy scenes, two key phenomena are observed: 1) the sky region often exhibits lower brightness compared to non-sky areas, contrasting with daytime conditions \cite{jin2023enhancing}; 2) the pixel intensity of an area is related to its depth, where the farther away from the camera, the lower the brightness may be \cite{liu2023nighthazeformer,sharma2020single}. Therefore, the illumination of daytime images cannot be adjusted in a global uniform way. The scene depth and sky area should be considered to make the synthesized low-light images match the two factors. To achieve this purpose, the depth estimation and sky segmentation ~\cite{depth_anything_v2} are used to adjust the brightness. The depth for the real-world daytime image $y_{d}^{r}$ is denoted as $g_{y_{d}^{r}}$. We define the concept of the non-uniform illumination mask $\mathcal{S}_{y_{d}^{r}}$, which is initialized by $g_{y_{d}^{r}}$. Then, $\mathcal{S}_{y_{d}^{r}}$ is split by the area where depth $g_{y_{d}^{r}}(i, j)$ is larger than the threshold $\varrho$ for the sky $v$. The intensity of $\mathcal{S}_{y_{d}^{r}}$ is adjusted as:
\begin{equation}
   \dot{\mathcal{S}}_{y_{d}^{r}}(i, j) = \left\{
   \begin{aligned}
     & \mathcal{S}_{y_{d}^{r}}(i, j) \odot \varphi_{1}, \hspace{0.2cm} 1 - g_{y_{d}^{r}}(i, j) \geq \varrho, \\
     & \mathcal{S}_{y_{d}^{r}}(i, j) \odot \varphi_{2}, \hspace{0.2cm} 1 - g_{y_{d}^{r}}(i, j) < \varrho, 
   \end{aligned}
   \right.
\end{equation}
where $i$ and $j$ denote the pixel locations. $\varphi_{1}$ and $\varphi_{2}$ represent the illumination factor for sky and non-sky areas, respectively. Further, we adjust the brightness of the sky area to make it lower than the foreground, as follows:
\begin{equation}
   \dot{y}_{d}^{r}(i, j) = \left\{
     \begin{aligned}
       & y_{d}^{r}(i, j)/\rho, \hspace{0.2cm} 1 - g_{y_{d}^{r}}(i, j) \geq \varrho, \\
       & y_{d}^{r}(i, j), \hspace{0.2cm} 1 - g_{y_{d}^{r}}(i, j) < \varrho, 
     \end{aligned}
     \right.
  \end{equation}
where $\rho$ is the brightness difference factor. We perform pixel-wise Gamma \cite{cong2024semi} to obtain $y_{n}^{r}$ by:
\begin{equation}
   \label{eq:the_simulated_nighttime_clear_image}
   {y}_{n}^{r}(i, j) = [\dot{y}_{d}^{r}(i, j)]^{\alpha \odot \dot{\mathcal{S}}_{y_{d}^{r}}(i, j)},
\end{equation}
where $\alpha$ is the degradation factor. Meanwhile, the brightness of the sky area is adjusted to make the mean value close to the preset level $\mu$. The above non-uniform brightness adjustment is one of the keys to our methods.

\noindent \textbf{Simulated distortion}. Since sufficiently distortions from the real world are not available, we synthesize them algorithmically. The farther away a point is from the light center, the smaller its brightness value is. Therefore, we follow the settings in the game engine \cite{liu2023nighthazeformer} and use the attenuation function to approximately represent the attenuation map $\mathcal{A}$ for the pixel $m$ as:
\begin{equation}
  \mathcal{A}(m, c) = [\xi_{1} + \xi_{2} \mathcal{D}(m, c) + \xi_{3} \mathcal{D}(m, c)^{2}]^{-1},
\end{equation}
where $\xi_{1}$, $\xi_{2}$, and $\xi_{3}$ are the hyperparameters that control the relationship between position and attenuation degree. $\mathcal{D}$ denotes the coordinate distance from $m$ to the light center $c$. Real-world active lighting devices include not only isotropic point light sources, but also directional cone-shaped light sources (such as street lamps with masks). To enrich the types of lighting regions, we approximately simulate point $\mathcal{I}^{P}$ and cone $\mathcal{I}^{C}$ light sources by: 
\begin{equation}
   \label{eq:light_centers}
  \left\{
    \begin{aligned}
      & \mathcal{I}^{P}(i_{m}, j_{m}) = \mathcal{A}(i_{m}, j_{m}) \odot \beta,\\
      & \mathcal{I}^{C}(i_{m}, j_{m}) = \mathcal{A}(i_{m}, j_{m}) \odot \phi(\cos{(\vec{a}, \vec{c})}) \odot \mathcal{Q}(\vec{a}, \vec{c}),
    \end{aligned}
  \right.
\end{equation}
where $\phi$ and $\beta$ are clip operation and constant. $\vec{a}$ and $\vec{c}$ represent the direction of pixel and the direction of cone light, respectively. $\mathcal{Q}$ is the area where light is irradiated. The color rendering \cite{zhang2020nighttime} is adopted to provide color information. Then, the rendered result is added to the Eq.~\ref{eq:the_simulated_nighttime_clear_image} to obtain the hazy image $x_{n}^{s}$ in Eq.~\ref{eq:severe_distortion_model}.

\begin{figure*}
   \centering
   \footnotesize

   \includegraphics[width=2cm,height=1.45cm]{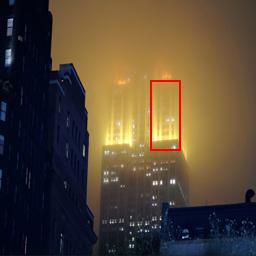}
   \includegraphics[width=0.6cm,height=1.45cm]{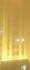}
   \includegraphics[width=0.1cm,height=1.5cm]{00_figures/plot_vertical_line/plot_vertical_line.png}
   \includegraphics[width=2cm,height=1.45cm]{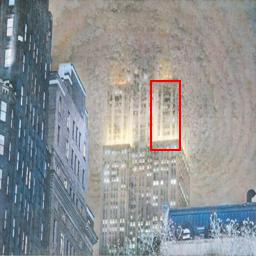}
   \includegraphics[width=0.6cm,height=1.45cm]{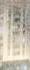}
   \hspace{0.1cm}
   \includegraphics[width=2cm,height=1.45cm]{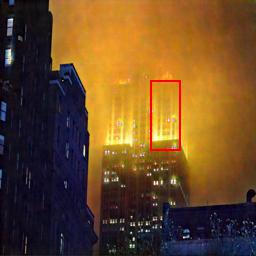}
   \includegraphics[width=0.6cm,height=1.45cm]{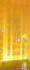}
   \hspace{0.1cm}
   \includegraphics[width=2cm,height=1.45cm]{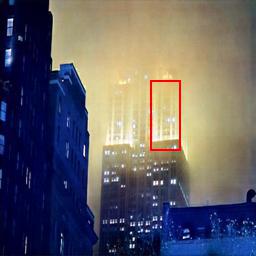}
   \includegraphics[width=0.6cm,height=1.45cm]{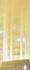}
   \hspace{0.1cm}
   \includegraphics[width=2cm,height=1.45cm]{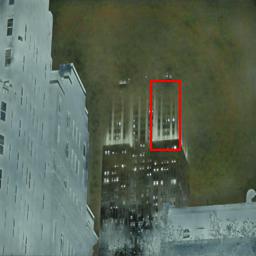}
   \includegraphics[width=0.6cm,height=1.45cm]{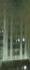}
   \includegraphics[width=0.1cm,height=1.5cm]{00_figures/plot_vertical_line/plot_vertical_line.png}
   \includegraphics[width=2cm,height=1.45cm]{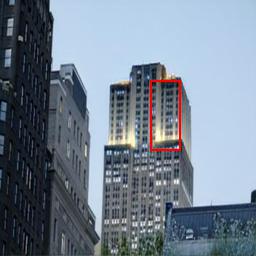}
   \includegraphics[width=0.6cm,height=1.45cm]{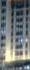}


   \includegraphics[width=2cm,height=1.45cm]{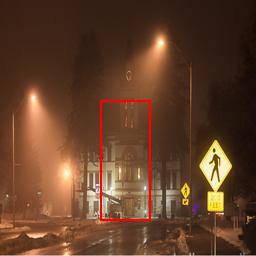}
   \includegraphics[width=0.6cm,height=1.45cm]{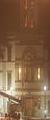}
   \includegraphics[width=0.1cm,height=1.5cm]{00_figures/plot_vertical_line/plot_vertical_line.png}
   \includegraphics[width=2cm,height=1.45cm]{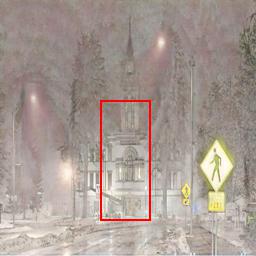}
   \includegraphics[width=0.6cm,height=1.45cm]{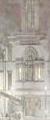}
   \hspace{0.1cm}
   \includegraphics[width=2cm,height=1.45cm]{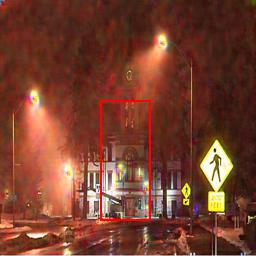}
   \includegraphics[width=0.6cm,height=1.45cm]{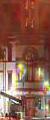}
   \hspace{0.1cm}
   \includegraphics[width=2cm,height=1.45cm]{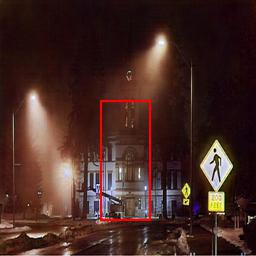}
   \includegraphics[width=0.6cm,height=1.45cm]{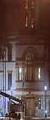}
   \hspace{0.1cm}
   \includegraphics[width=2cm,height=1.45cm]{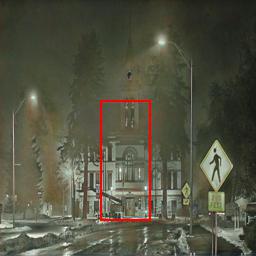}
   \includegraphics[width=0.6cm,height=1.45cm]{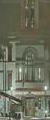}
   \includegraphics[width=0.1cm,height=1.5cm]{00_figures/plot_vertical_line/plot_vertical_line.png}
   \includegraphics[width=2cm,height=1.45cm]{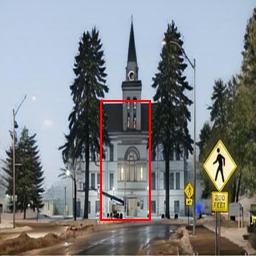}
   \includegraphics[width=0.6cm,height=1.45cm]{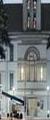}


   \includegraphics[width=2cm,height=1.45cm]{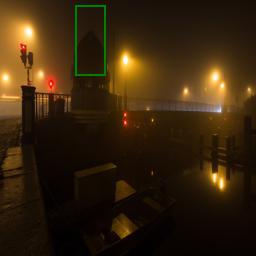}
   \includegraphics[width=0.6cm,height=1.45cm]{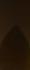}
   \includegraphics[width=0.1cm,height=1.5cm]{00_figures/plot_vertical_line/plot_vertical_line.png}
   \includegraphics[width=2cm,height=1.45cm]{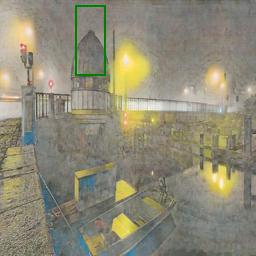}
   \includegraphics[width=0.6cm,height=1.45cm]{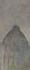}
   \hspace{0.1cm}
   \includegraphics[width=2cm,height=1.45cm]{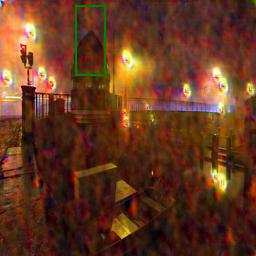}
   \includegraphics[width=0.6cm,height=1.45cm]{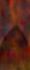}
   \hspace{0.1cm}
   \includegraphics[width=2cm,height=1.45cm]{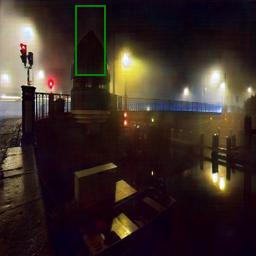}
   \includegraphics[width=0.6cm,height=1.45cm]{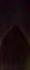}
   \hspace{0.1cm}
   \includegraphics[width=2cm,height=1.45cm]{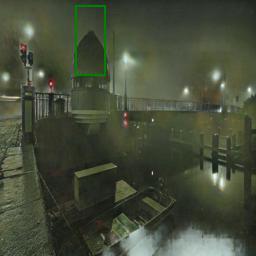}
   \includegraphics[width=0.6cm,height=1.45cm]{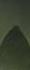}
   \includegraphics[width=0.1cm,height=1.5cm]{00_figures/plot_vertical_line/plot_vertical_line.png}
   \includegraphics[width=2cm,height=1.45cm]{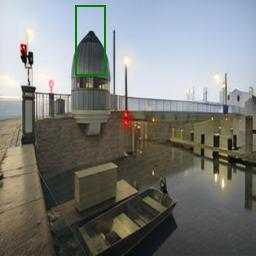}
   \includegraphics[width=0.6cm,height=1.45cm]{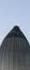}


   \includegraphics[width=2cm,height=1.45cm]{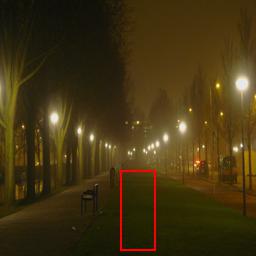}
   \includegraphics[width=0.6cm,height=1.45cm]{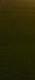}
   \includegraphics[width=0.1cm,height=1.5cm]{00_figures/plot_vertical_line/plot_vertical_line.png}
   \includegraphics[width=2cm,height=1.45cm]{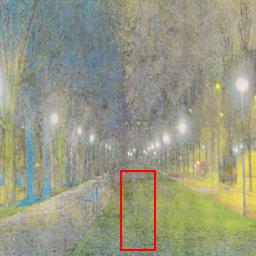}
   \includegraphics[width=0.6cm,height=1.45cm]{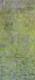}
   \hspace{0.1cm}
   \includegraphics[width=2cm,height=1.45cm]{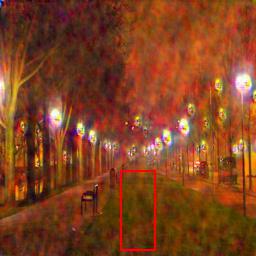}
   \includegraphics[width=0.6cm,height=1.45cm]{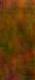}
   \hspace{0.1cm}
   \includegraphics[width=2cm,height=1.45cm]{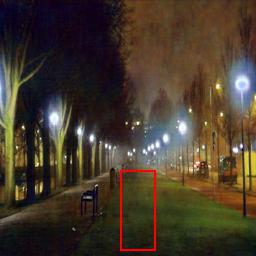}
   \includegraphics[width=0.6cm,height=1.45cm]{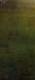}
   \hspace{0.1cm}
   \includegraphics[width=2cm,height=1.45cm]{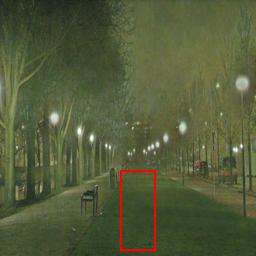}
   \includegraphics[width=0.6cm,height=1.45cm]{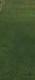}
   \includegraphics[width=0.1cm,height=1.5cm]{00_figures/plot_vertical_line/plot_vertical_line.png}
   \includegraphics[width=2cm,height=1.45cm]{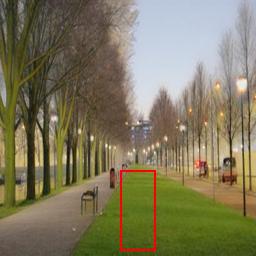}
   \includegraphics[width=0.6cm,height=1.45cm]{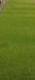}


   \includegraphics[width=2cm,height=1.45cm]{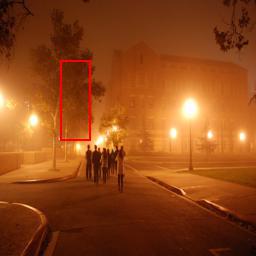}
   \includegraphics[width=0.6cm,height=1.45cm]{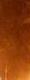}
   \includegraphics[width=0.1cm,height=1.5cm]{00_figures/plot_vertical_line/plot_vertical_line.png}
   \includegraphics[width=2cm,height=1.45cm]{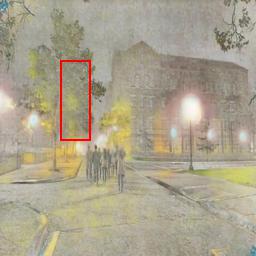}
   \includegraphics[width=0.6cm,height=1.45cm]{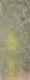}
   \hspace{0.1cm}
   \includegraphics[width=2cm,height=1.45cm]{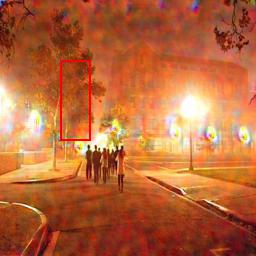}
   \includegraphics[width=0.6cm,height=1.45cm]{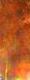}
   \hspace{0.1cm}
   \includegraphics[width=2cm,height=1.45cm]{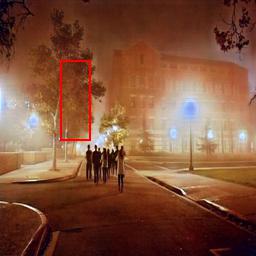}
   \includegraphics[width=0.6cm,height=1.45cm]{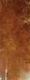}
   \hspace{0.1cm}
   \includegraphics[width=2cm,height=1.45cm]{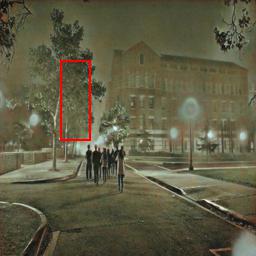}
   \includegraphics[width=0.6cm,height=1.45cm]{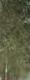}
   \includegraphics[width=0.1cm,height=1.5cm]{00_figures/plot_vertical_line/plot_vertical_line.png}
   \includegraphics[width=2cm,height=1.45cm]{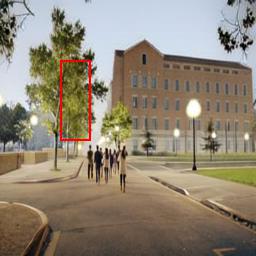}
   \includegraphics[width=0.6cm,height=1.45cm]{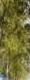}

   \leftline{\hspace{1cm} (a) Hazy \hspace{1cm} (b) by UNREAL-NH \hspace{0.4cm} (c) by NHR-Daytime \hspace{0.3cm} (d) by NHM-Daytime \hspace{0.4cm} (e) by NHC-Daytime \hspace{0.6cm} (f) Our method}
   \caption{Real-world visual comparisons with existing NDBM-Task solutions.}
   \label{fig:comparison_with_NDE_Task_solutions}
 \end{figure*}

\subsection{Diffusion Model for Lighting Reconstruction}
To achieve daytime lighting reconstruction, we propose to train the diffusion model under the guidance of a brightness perception network.

\noindent \textbf{Preliminary.} The training objective of the diffusion model in the latent space \cite{rombach2022high} is:
\begin{equation}
   \mathcal{L} = \mathbb{E}_{\mathcal{H}_{e}(x), p, \epsilon \sim \mathcal{N}(0, 1), t }\Big[ \Vert \epsilon - \epsilon_{\theta}(z_{t},t, \tau_{\theta}(p)) \Vert_{2}^{2}\Big] \, ,
   \label{eq:cond_loss}
\end{equation}
where $z_{t}$ and $\tau_{\theta}$ are latent variable and domain specific encoder \cite{rombach2022high}, respectively. In our NDBM-Task, we need to reconstruct a high-quality daytime image. Models with prior knowledge can better understand the distribution of daytime images, thereby improving image reconstruction performance. To achieve this purpose, we adopt the one-step diffusion \cite{parmar2024one,sauer2024adversarial} as a backbone model for injecting conditional information (hazy images) into the pre-trained model \cite{parmar2024one}. Similar to existing solutions \cite{parmar2024one}, LoRA \cite{hu2021lora} is used to fine-tune the $\mathcal{H}_{e}$, $\mathcal{H}_{u}$, and $\mathcal{H}_{k}$.

\begin{table*}
   \footnotesize
   \centering
   \setlength{\tabcolsep}{2.5mm}
   \renewcommand{\arraystretch}{1.2}
   \caption{Comparison of the no-reference metrics with NDBM-Task solutions. The best performances are in bold.}
   \label{tab:NR_metrics_compare_with_NDE_solutions}
   
   \begin{tabular}{ccccccccccc}
     \hline

     \hline   
                        
                        & Q-ALIGN  & QualiCLIP+ & LIQE  & ARNIQA  & TOPIQ  & TReS  & CLIPIQA  & MANIQA  & MUSIQ  & DBCNN  \\
                        
    \hline
     NHR-Daytime    & 1.992 & 0.406 & 1.403 & 0.494 &  0.342 & 45.221 &  0.326 &  0.227 &  40.284 & 0.359 \\
                  
     NHM-Daytime    &  2.321 &  0.427 & 1.596 & 0.548 &  0.360 & 48.922 & 0.401 & 0.222 &  42.714 &  0.371 \\
                  
     NHC-Daytime     &  2.001 & 0.449 &  1.300 &  0.557 &  0.374 &  58.957 &  0.352 &  0.246 &  45.643 &  0.408  \\
                  
     UNREAL-NH      &   1.795 & 0.432 &  1.249 &  0.549 & 0.356 &  56.691 &  0.314 &  0.227 &  43.285 &  0.378 \\
                  
    \hline
    Ours            & \textbf{2.539} & \textbf{0.573} &  \textbf{2.853} & \textbf{0.640} &  \textbf{0.443} &  \textbf{61.050} & \textbf{0.420} & \textbf{0.297} & \textbf{51.098} &  \textbf{0.414} \\
   \hline
   \end{tabular}
 \end{table*}

\begin{figure*}
   \centering
   \footnotesize

   \includegraphics[width=1.62cm,height=1.6cm]{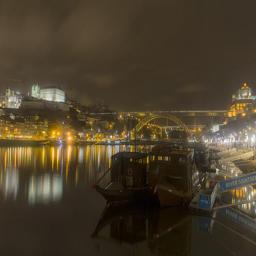}
   \includegraphics[width=1.62cm,height=1.6cm]{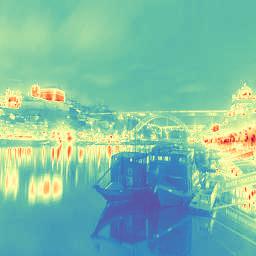}
   \includegraphics[width=0.1cm,height=1.5cm]{00_figures/plot_vertical_line/plot_vertical_line.png}
   \includegraphics[width=1.62cm,height=1.6cm]{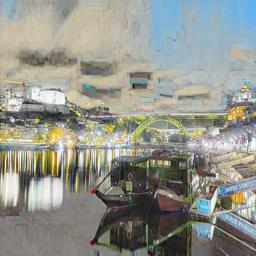}
   \includegraphics[width=1.62cm,height=1.6cm]{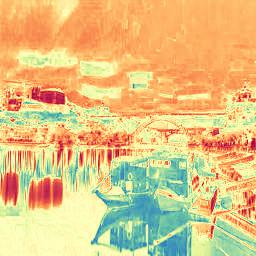}
   \hspace{0.1cm}
   \includegraphics[width=1.62cm,height=1.6cm]{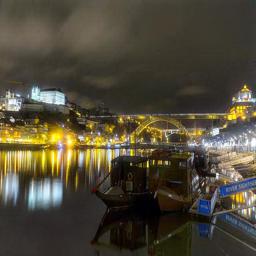}
   \includegraphics[width=1.62cm,height=1.6cm]{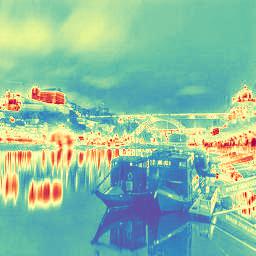}
   \hspace{0.1cm}
   \includegraphics[width=1.62cm,height=1.6cm]{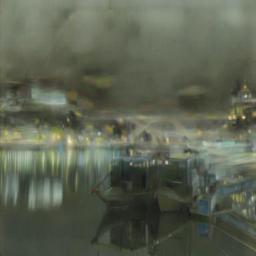}
   \includegraphics[width=1.62cm,height=1.6cm]{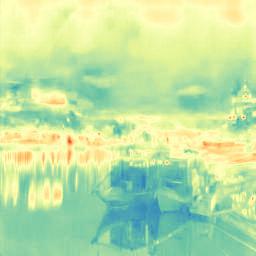}
   \includegraphics[width=0.1cm,height=1.5cm]{00_figures/plot_vertical_line/plot_vertical_line.png}
   \includegraphics[width=1.62cm,height=1.6cm]{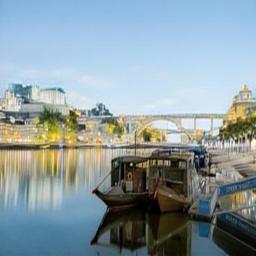}
   \includegraphics[width=1.62cm,height=1.6cm]{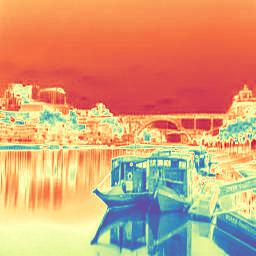}

   \leftline{\hspace{1.1cm} (a) Hazy \hspace{1.7cm} (b) on UNREAL-NH \hspace{1cm} (c) on NHR-Daytime \hspace{1cm} (d) on NHC-Daytime \hspace{1cm} (e) on our dataset}
   \caption{Real-world visual comparisons of the results obtained by training our model under different datasets.}
   \label{fig:train_our_model_on_existing_datasets}
 \end{figure*}

\begin{figure*}[!tb]
   \centering
   \footnotesize


   \includegraphics[width=1.6cm,height=1.6cm]{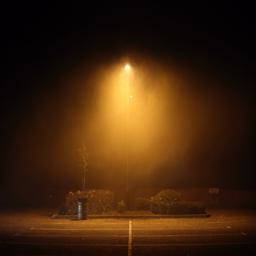}
   \includegraphics[width=1.6cm,height=1.6cm]{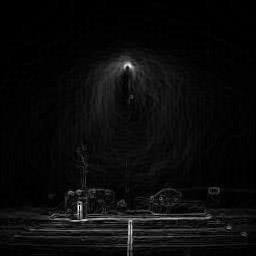}
   \includegraphics[width=0.1cm,height=1.4cm]{00_figures/plot_vertical_line/plot_vertical_line.png}
   \includegraphics[width=1.6cm,height=1.6cm]{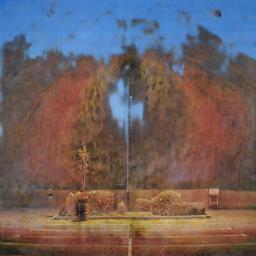}
   \includegraphics[width=1.6cm,height=1.6cm]{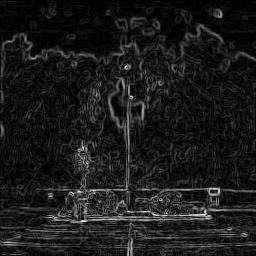}
   \hspace{0.12cm}
   \includegraphics[width=1.6cm,height=1.6cm]{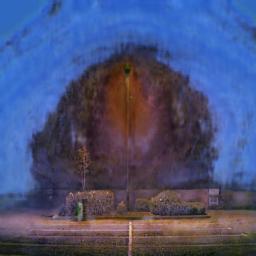}
   \includegraphics[width=1.6cm,height=1.6cm]{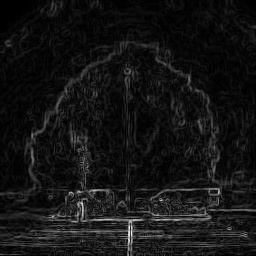}
   \hspace{0.12cm}
   \includegraphics[width=1.6cm,height=1.6cm]{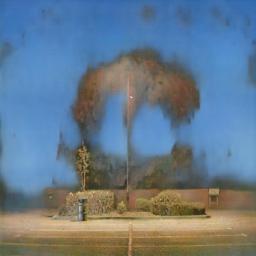}
   \includegraphics[width=1.6cm,height=1.6cm]{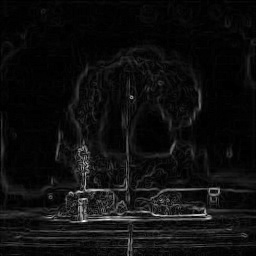}
   \includegraphics[width=0.1cm,height=1.4cm]{00_figures/plot_vertical_line/plot_vertical_line.png}
   \includegraphics[width=1.6cm,height=1.6cm]{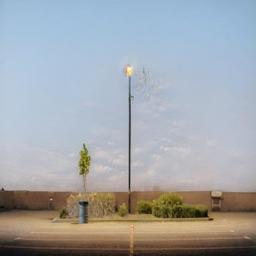}
   \includegraphics[width=1.6cm,height=1.6cm]{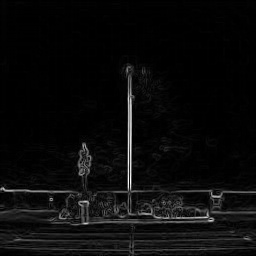}

   \leftline{\hspace{1.0cm} (a) Hazy \hspace{2cm} (b) by Fourmer \hspace{2cm} (c) by GAC \hspace{1.8cm} (d) by SFSNiD \hspace{1.7cm} (e) by our model}
   \caption{Real-world visual comparisons of the results obtained by training different models under our dataset. The edge maps obtained by the Sobel operation \cite{zhou2022edge} indicate our model can obtain clear structures.}
   \label{fig:train_our_dataset_on_existing_models}
 \end{figure*}

\noindent \textbf{Training stage.} 
Since the brightness of nighttime hazy scenes is non-uniform, we embed this property into the optimization process of the diffusion model as shown in Fig.~\ref{fig:data_synthesis_network_training_and_inference_process}. The brightness perception network is denoted as $\Phi$, which is implemented by naive convolutions. The feature layers extracted from the encoder $\mathcal{H}_{e}$ and decoder $\mathcal{H}_{k}$ are denoted as $h_{e}^{z}$ and $h_{k}^{z}$, where $z \in \{1, 2, 3, 4\}$. The $\Phi$ uses the first layer features of $\mathcal{H}_{e}$ as input. The predict brightness map $\hat{\mathcal{M}}$ for $x_{n}^{s}$ is: 
\begin{equation}
   \hat{\mathcal{M}} = \Phi(h_{e}^{1}).
\end{equation}

In the image reconstruction task \cite{zhang2020nighttime,parmar2023zero}, passing information from the encoder to the decoder by skip layer connections has been shown to effectively improve the structural restoration ability. Therefore, we use the feature extraction module $\Gamma^{z}(\cdot)$ in \cite{cong2024semi} to pass the encoder features to the decoder. During the feature transfer process, the features of the skip connections will perceive the brightness map from $\hat{\mathcal{M}}$. The feature maps after skip connections are:
\begin{equation}
   \left\{
      \begin{aligned}
        & \hat{h}_{k}^{1} = \Gamma^{1}(h_{e}^{1}) \odot (1 + \hat{\mathcal{M}}) + h_{k}^{1}, \\
        & \hat{h}_{k}^{z} = \Gamma^{z}(h_{e}^{z}) + h_{k}^{z}, z=2,3,4,
      \end{aligned}
      \right.
\end{equation}
where the weighting process of the brightness information as $f(h_{e}^{1}, \hat{\mathcal{M}}) = \Gamma^{1}(h_{e}^{1}) \odot (1 + \hat{\mathcal{M}})$ as shown in Fig~\ref{fig:data_synthesis_network_training_and_inference_process}. The brightness mean is used to define the reference label $\mathcal{M}$. The loss for the brightness information is: 
\begin{equation}
   \mathcal{L}_{\mathcal{M}} = \sum_{o=1}^{\mathcal{O}} (\hat{\mathcal{M}}[o] - \mathcal{M}[o])^{2},
\end{equation}
where $\mathcal{O}$ is the total number of examples. During the training stage, we adopt a fixed daytime prompt $p_{d}$ ``a clear daytime image''. Since a strict pixel-wise correspondence is needed in our task, a pixel-by-pixel constraint is used as:
\begin{equation}
   \mathcal{L}_{pc} = \sum_{o=1}^{\mathcal{O}} |\mathcal{H}_{k}(\mathcal{H}_{u}(\tau_{\theta}(p_{d}), \mathcal{H}_{e}(x_{n}^{s}[o]))) - y_{d}^{r}[o]|.
\end{equation}
To further improve the realism of generated images, we adopt adversarial loss \cite{parmar2024one,kumari2022ensembling} to discriminate generated images from real images as:
\begin{equation}
     \begin{aligned}
      \mathcal{L}_{adv} =  & \mathbb{E}_{y_{d}^{r}}\log(\mathcal{R}(y_{d}^{r})) + \\
       & \mathbb{E}_{x_{n}^{s}}\log(1 - \mathcal{R}(\mathcal{H}_{k}(\mathcal{H}_{u}(\tau_{\theta}(p_{d}), \mathcal{H}_{e}(x_{n}^{s}))))),
     \end{aligned}
\end{equation}
where $\mathcal{R}$ is the discriminator. The overall loss function is:
\begin{equation}
   \mathcal{L} = \mathcal{L}_{pc} + \lambda_1 \mathcal{L}_{adv} + \lambda_2 \mathcal{L}_{\mathcal{M}},
\end{equation}
where $\lambda_1$ and $\lambda_2$ are weight factors. 

\noindent \textbf{Inference stage.} By using the daytime prompt $p_{d}$, we can obtain a clear image in daytime brightness for the real-world hazy image $x_{n}^{r}$, which is:
\begin{equation}
   \hat{y}_{d}^{r} = \mathcal{H}_{k}(\mathcal{H}_{u}(\tau_{\theta}(p_{d}), \mathcal{H}_{e}(x_{n}^{r}))).
\end{equation}

Notably, although the purpose of this paper is to obtain results with daytime brightness, our model also can generate nighttime results. During the training stage, we only fine-tune part of the layers, which preserves the semantic understanding ability. Therefore, we can use ``daytime'' and ``nighttime'' as different control information. By using the nighttime prompt $p_{n}$ ``a dark nighttime image'', we can obtain a dehazed image with nighttime illuminations.


\section{Experiments}
\subsection{Settings}
\noindent \textbf{Our dataset.} We manually select our training data from Places365 \cite{zhou2017places} datadset. The training data include 1546 images with blue sky. Our test data (440 images) comes from the real-world nighttime haze (RWNH) \cite{jin2023enhancing}.

\noindent \textbf{Comparison.} The used datasets include: (1) NH-series (NHR-Daytime, NHM-Daytime, NHC-Daytime) \cite{zhang2020nighttime}, (2) UNREAL-NH \cite{liu2023nighthazeformer}. The NH-series datasets include both daytime and nighttime labels, we use the suffixes ``Daytime'' to show that we are using the daytime labels. Currently, models designed for NDBM-Task are scarce. Since neural networks have strong data fitting capabilities, we use the models that verified in dehazing and general restoration tasks, including Fourmer \cite{zhou2023fourmer}, GAC \cite{jin2023enhancing} and SFSNiD\cite{cong2024semi}, which have advanced performance \cite{cong2024semi}.

\noindent \textbf{Evaluations.} Quantitative evaluation metrics \cite{wu2024comprehensive,pyiqa,wei2024robust,wei2024robust_2} including QualiCLIP \cite{agnolucci2024quality}, ARNIQA \cite{agnolucci2024arniqa}, TOPIQ \cite{chen2024topiq}, MUSIQ \cite{ke2021musiq}, DBCNN \cite{zhang2020blind}, Q-ALIGN \cite{wu2023q}, LIQE \cite{zhang2023blind}, TReS \cite{golestaneh2021no}, CLIPIQA \cite{wang2022exploring}, and MANIQA \cite{yang2022maniqa}. For the sake of intuition, six metrics (QualiCLIP+, ARNIQA, TOPIQ, CLIPIQA, MANIQA, DBCNN) with similar ranges are used for plotting, abbreviated as NR-1 to NR-6.

\noindent \textbf{Implementation details.} We use $512 \times 512$ size for training. The experimental platform is a single NVIDIA RTX 4090. Adam optimizer is used. The batch size is set to 1. The ranks of LoRA for UNet and $\{$encoder, decoder$\}$ are 8 and 4, respectively. $\lambda_1$ and $\lambda_2$ are set to 0.5 and 1, respectively. The sky region segmentation parameter $\varrho$ is empirically set to 0.98. The $\mu$ is 0.85. $\varphi_{1}$ and $\varphi_{2}$ are $2$ and $1.5$, respectively. $\alpha$ is set to 4. $\beta$ is set to 1. $\xi_{1}$, $\xi_{2}$, and $\xi_{3}$ are set to 1, 3, and 1.8, respectively.

\subsection{Comparison with Existing Solutions}
Fig.~\ref{fig:comparison_with_NDE_Task_solutions} shows the dehazing performance obtained on existing datasets under the state-of-the-art dehazing model \cite{cong2024semi}. Obviously, existing solutions cannot achieve satisfactory results. Neither the brightness of the image nor the content of the scene is noticeably improved. The brightness of the image obtained by UNREAL-NH is too high. The images obtained by NH-series cannot complete the night-to-day brightness mapping.

The visual results show that the results obtained by our method are closest to real-world daytime brightness, which verifies the data synthesis pipeline can achieve night-to-day mapping. The content visibility of the scene is overall improved. We provide visual results by comparing with more dehazing models under these datasets in the Supplementary Materials, and the results are the same. Furthermore, Table~\ref{tab:NR_metrics_compare_with_NDE_solutions} verifies that the proposed method achieves better no-reference evaluation results. Visual and quantitative results demonstrate that our approach can obtain better haze removal and brightness mapping performance.

\begin{figure}
   \footnotesize
   \centering
   \includegraphics[width=\linewidth]{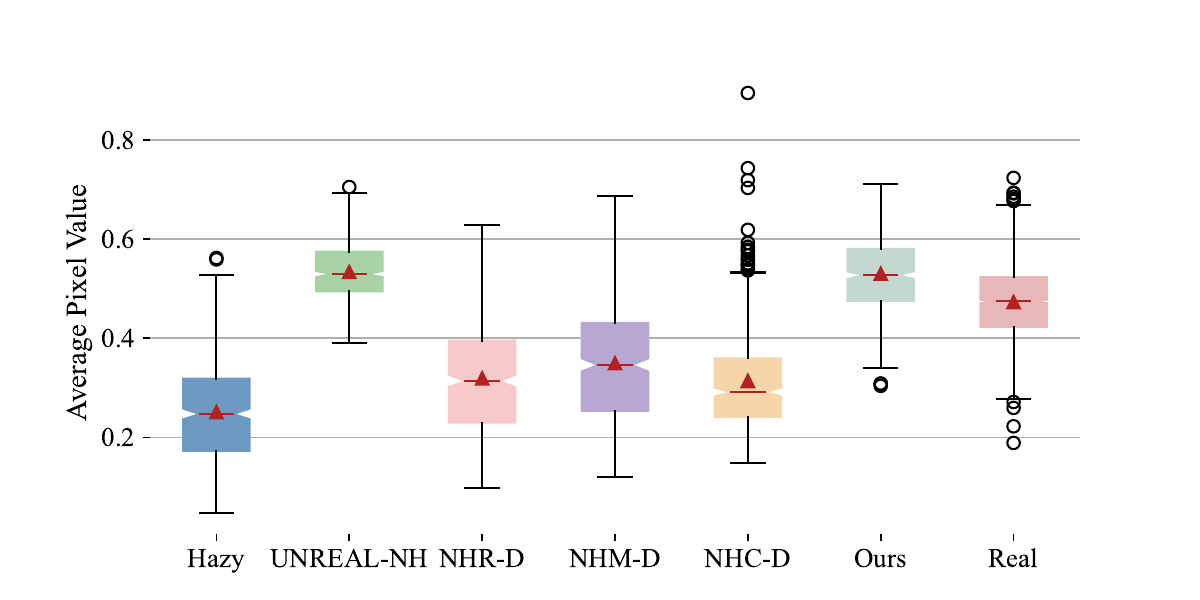}
   \caption{Brightness statistics obtained from training our model with different datasets. ``-D'' means Daytime. The ``Real'' comes from the real-world clear images in OTS \cite{li2018benchmarking}.}
   \label{fig:brightness_statistics_obtained_from_training_our_model_with_different_datasets}
\end{figure}

\begin{table*}
   \footnotesize
   \centering
   \setlength{\tabcolsep}{2.3mm}
   \caption{Comparison of metrics obtained by training our model using existing datasets. The best performances are in bold.}
   \label{tab:train_our_model_on_existing_daytime_datasets}
   
   \begin{tabular}{ccccccccccc}
     \hline

     \hline   
                        & Q-ALIGN  & QualiCLIP+ & LIQE  & ARNIQA  & TOPIQ  & TReS  & CLIPIQA  & MANIQA  & MUSIQ  & DBCNN  \\
    \hline
     SD-NHR-Daytime     & 2.463 & 0.489 & 1.988 & 0.563 & 0.416 & 59.630 & \textbf{0.438} &  0.276 &  47.001 &  0.421 \\

     SD-NHM-Daytime     & 2.397 & 0.478 & 1.981 & 0.587 & 0.437 & 65.271 & 0.425 & 0.294 & 47.838 & \textbf{0.445} \\
     SD-NHC-Daytime     &  1.460 & 0.345 & 1.017 & 0.407 & 0.215 & 26.830 & 0.208 & 0.129 & 28.039 &  0.215 \\

     SD-UNREAL-NH      &  1.903 &  0.504 &  1.871 &  0.571 &  0.410 &  \textbf{68.299} &  0.278 &  0.283 & \textbf{52.881} & 0.426  \\
    \hline
    Ours               & \textbf{2.539} & \textbf{0.573} &  \textbf{2.853} & \textbf{0.640} &  \textbf{0.443} & 61.050 & 0.420 & \textbf{0.297} & 51.098 &  0.414 \\ 
   \hline
   \end{tabular}
 \end{table*}

\begin{table*}
   \footnotesize
   \centering
   \setlength{\tabcolsep}{2.8mm}
   \caption{Comparison of metrics obtained by training our dataset using existing models. The best performances are in bold.}
   \label{tab:train_our_dataset_by_existing_models}
   
   \begin{tabular}{ccccccccccc}
     \hline

     \hline   
                        & Q-ALIGN  & QualiCLIP+ & LIQE  & ARNIQA  & TOPIQ  & TReS  & CLIPIQA  & MANIQA  & MUSIQ  & DBCNN  \\
    \hline
     GAC                & 1.947 &  0.400 & 1.346 & 0.575 & 0.356 & 50.480 & 0.284 & 0.203 & 44.689 & 0.342 \\
                        
     Fourmer            & 2.299 & 0.469 &  1.776 & 0.616 & \textbf{0.449} & \textbf{64.665} & 0.345 & 0.270 & 50.793 & \textbf{0.427} \\
                       
     SFSNiD             & 2.231  & 0.478 & 1.645 & 0.594 & 0.442 & 63.258 & 0.352 & 0.258 & 49.572 & 0.423   \\
    \hline
    Ours               & \textbf{2.539} & \textbf{0.573} &  \textbf{2.853} & \textbf{0.640} &  0.443 & 61.050 & \textbf{0.420} & \textbf{0.297} & \textbf{51.098} & 0.414 \\

   \hline
   \end{tabular}
 \end{table*}

\begin{figure}
   \centering
   \footnotesize
   \includegraphics[width=5.8cm,height=0.4cm]{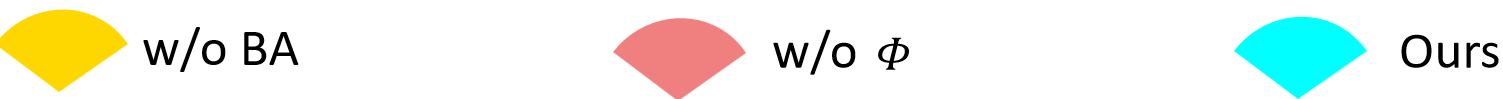}

   \includegraphics[width=1.5cm,height=1.5cm]{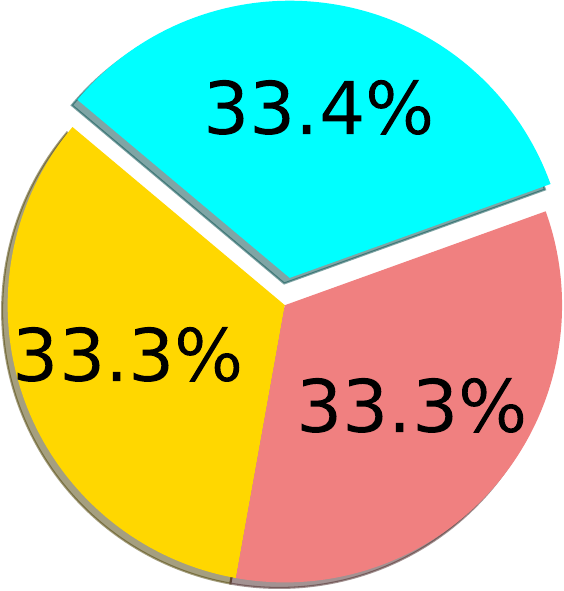}
   \includegraphics[width=1.5cm,height=1.5cm]{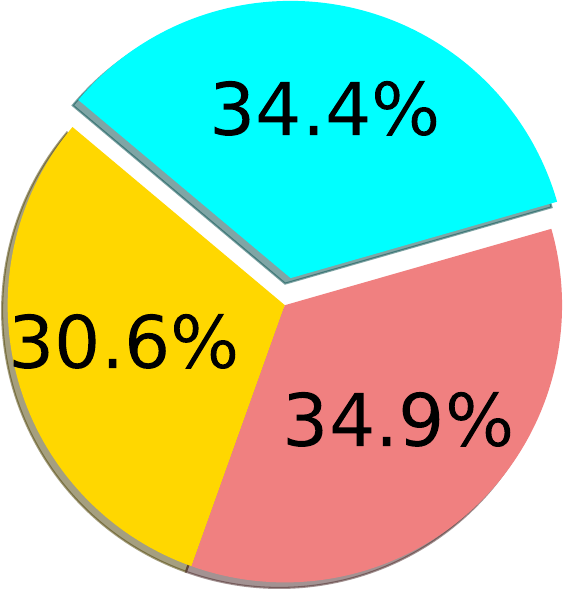}
   \includegraphics[width=1.5cm,height=1.5cm]{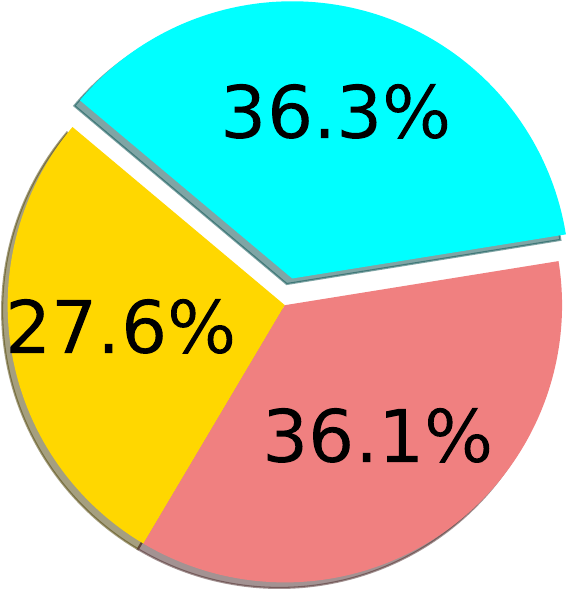}
   \includegraphics[width=1.5cm,height=1.5cm]{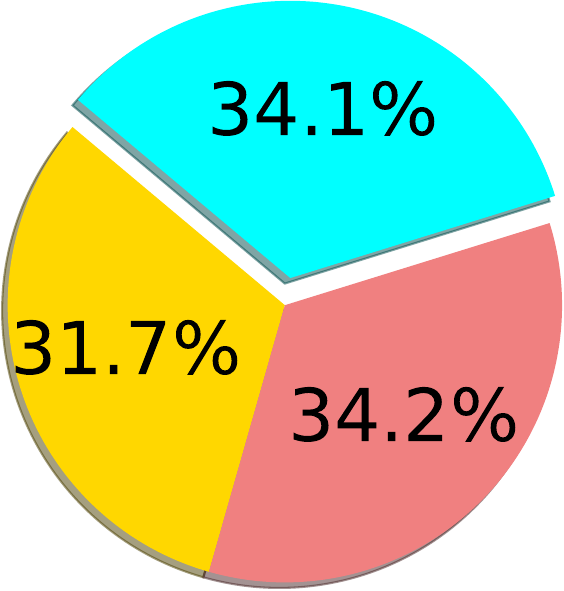}
   \includegraphics[width=1.5cm,height=1.5cm]{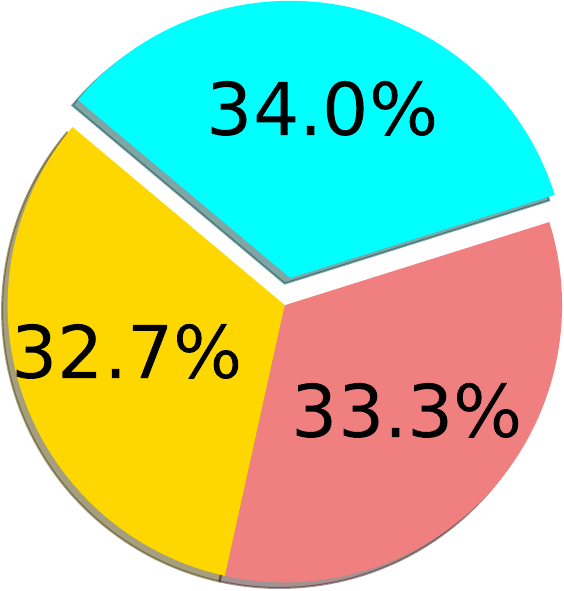}
   \leftline{\hspace{0.4cm} Q-ALIGN \hspace{0.2cm} QualiCLIP \hspace{0.3cm} ARNIQA \hspace{0.5cm} LIQE \hspace{0.7cm} TOPIQ}

   \includegraphics[width=1.5cm,height=1.5cm]{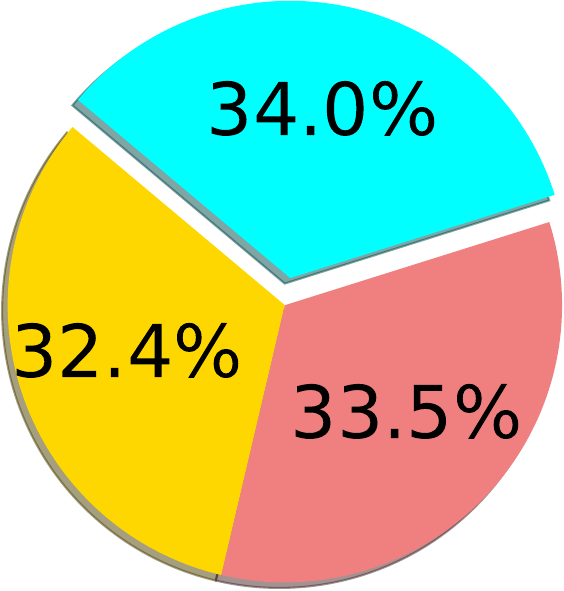}
   \includegraphics[width=1.5cm,height=1.5cm]{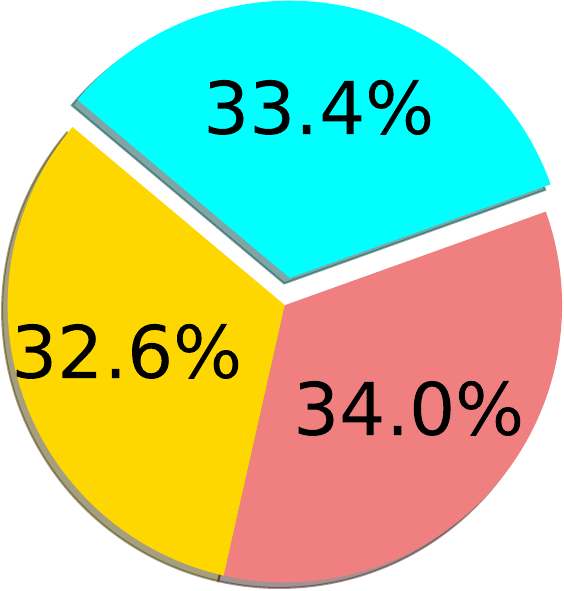}
   \includegraphics[width=1.5cm,height=1.5cm]{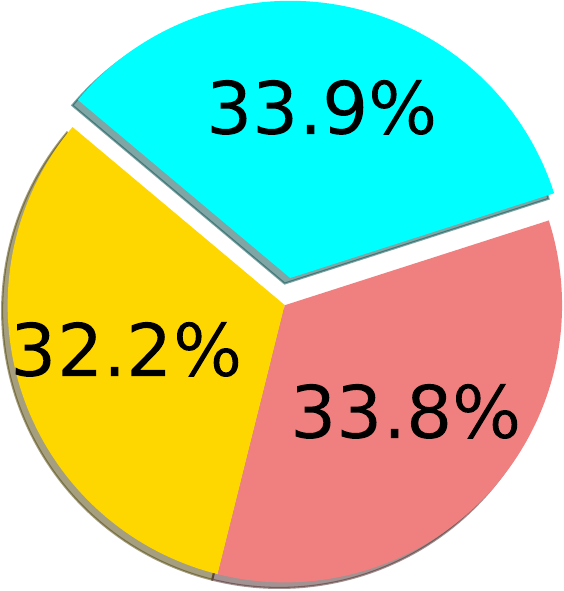}
   \includegraphics[width=1.5cm,height=1.5cm]{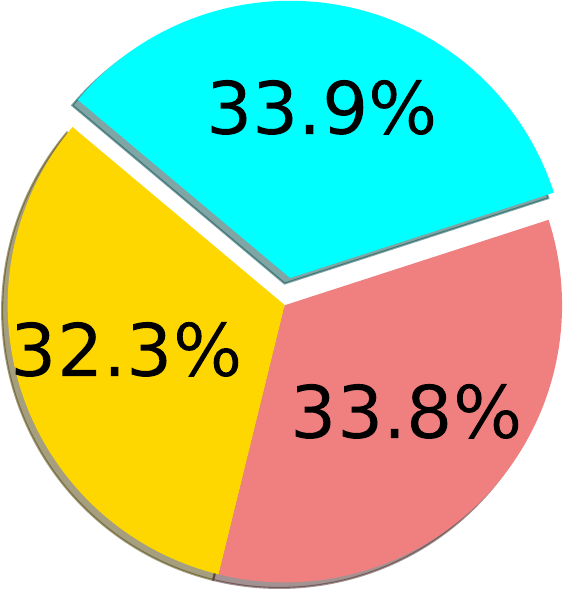}
   \includegraphics[width=1.5cm,height=1.5cm]{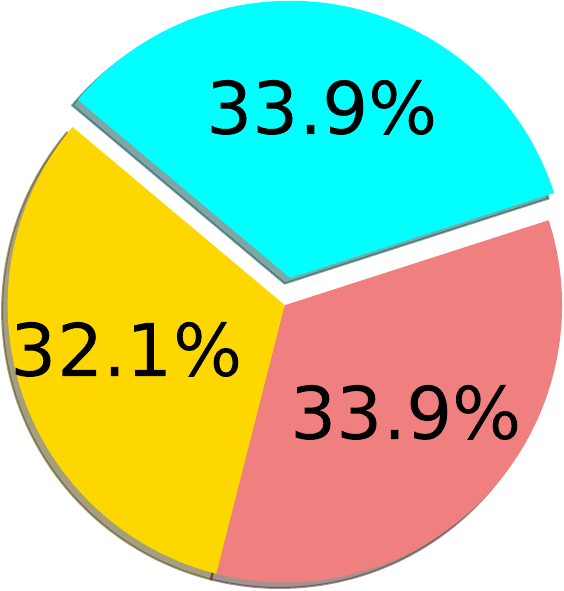}

   \leftline{\hspace{0.5cm} TReS \hspace{0.6cm} CLIPIQA \hspace{0.4cm} MANIQA \hspace{0.4cm} MUSIQ \hspace{0.45cm} DBCNN}

   \caption{Quantitative evaluation of the brightness adjustment (BA) and brightness perception network $\Phi$. The results denote the percentage of the metrics obtained by adding up the three settings.}
   \label{fig:quantitative_evaluation_of_the_sky_adjust_and_BrightnessPredNet}
 \end{figure}

\begin{figure}[!tb]
   \centering
   \footnotesize
   \includegraphics[width=2.0cm,height=1.4cm]{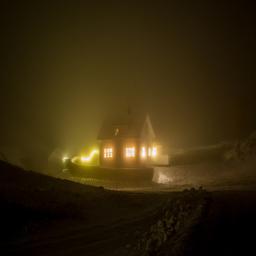}
   \includegraphics[width=2.0cm,height=1.4cm]{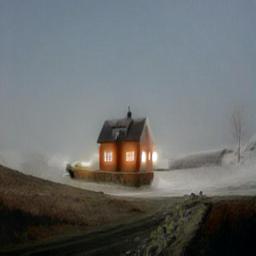}
   \includegraphics[width=2.0cm,height=1.4cm]{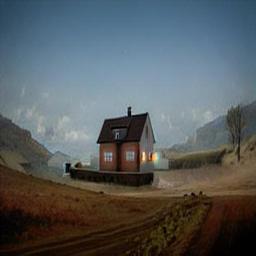}
   \includegraphics[width=2.0cm,height=1.4cm]{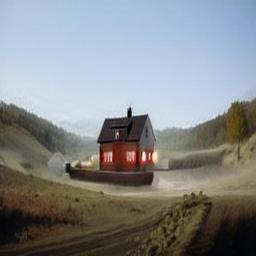}

   \includegraphics[width=2.0cm,height=1.4cm]{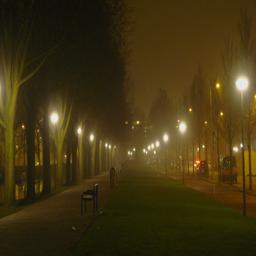}
   \includegraphics[width=2.0cm,height=1.4cm]{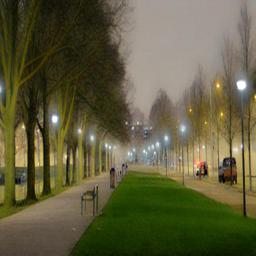}
   \includegraphics[width=2.0cm,height=1.4cm]{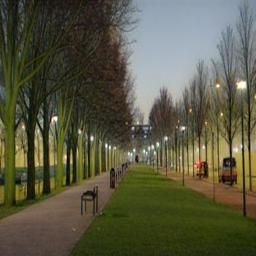}
   \includegraphics[width=2.0cm,height=1.4cm]{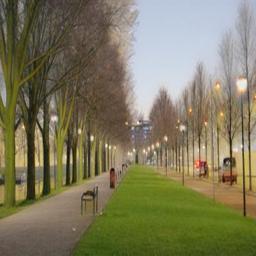}

   \leftline{\hspace{0.6cm} (a) Hazy \hspace{0.7cm} (b) w/o BA \hspace{0.7cm} (c) w/o $\Phi$ \hspace{0.9cm} (d) Ours}
   \caption{Visual evaluation of the brightness adjust (BA) and brightness perception network $\Phi$.}
   \label{fig:visual_evaluation_of_brightness_adjustment_and_BrightnessPredNet}
 \end{figure}

\subsection{Analysis of Our Dataset and Model}
\label{subsec:ablation_studies_of_our_dataset_and_model}
Our work consists of two aspects: data that enables brightness mapping, and a diffusion model guided by a brightness perception network for lighting reconstruction. Here, we conduct a cross-validation study to demonstrate the necessity of both.

\noindent \textbf{Train our model with existing datasets.} We use NHR-Daytime, NHC-Daytime, and UNREAL-NH for the training of our model. The results in Fig.~\ref{fig:train_our_model_on_existing_datasets} show that despite using our model, existing data synthesis strategies are still unable to achieve high-quality night-to-day brightness mapping. The dehazed images they obtain are not similar to real-world images. We count the brightness information and display it in Fig.~\ref{fig:brightness_statistics_obtained_from_training_our_model_with_different_datasets}.  The image brightness levels obtained by the UNREAL-NH and ours are close to the real world, but UNREAL-NH can not achieve realistic brightness as we proved. The brightness level obtained by the NH-series is too low. Meanwhile, the results in Table~\ref{tab:train_our_model_on_existing_daytime_datasets} show that using existing datasets to train our model cannot obtain the same metrics as ours. Visual and quantitative results prove the necessity of the dataset we designed.

\noindent \textbf{Train existing models on our dataset.} Models \cite{cong2024semi,jin2023enhancing,zhou2023fourmer} with sufficient data fitting ability are selected to train our proposed dataset. The results in Fig.~\ref{fig:train_our_dataset_on_existing_models} show that these methods are still unable to complete the lighting reconstruction. The content of the image is destroyed and the details of the object are lost. The possible reason is that they have not been trained on a large amount of real-world data, which limits their generalization ability to fit the lighting distribution of the real world. Meanwhile, Table~\ref{tab:train_our_dataset_by_existing_models} shows that the existing models cannot achieve effective dehazing ability in terms of quantitative metrics. Visual and quantitative results prove the necessity of the model we proposed.

\noindent \textbf{Overall conclusion.} The cross-validation results show that our data and model must be used together. Using either one alone will not produce visually friendly dehazing and brightness mapping results. Meanwhile, these two experiments prove that our analysis in Section \ref{sec:introduction} is reasonable.

\subsection{Ablation Studies and Discussions}

\noindent \textbf{Non-uniform brightness adjustment.} In order to study the necessity of the non-uniform brightness adjustment in data simulation, we remove this process and conduct quantitative and visual analysis. The results in Fig.~\ref{fig:quantitative_evaluation_of_the_sky_adjust_and_BrightnessPredNet} show that brightness adjustment has a certain impact on quantitative performance. Meanwhile, Fig. \ref{fig:visual_evaluation_of_brightness_adjustment_and_BrightnessPredNet} demonstrates that when the non-uniform brightness adjustment is removed, the brightness of the image is obviously reduced, which proves the necessity of the brightness adjustment we proposed.

\noindent \textbf{Brightness perception network $\Phi$ and $\mathcal{L}_{\mathcal{M}}$.}
We adopt a brightness perception network $\Phi$ during the training process, which is optimized under $\mathcal{L}_{\mathcal{M}}$. To demonstrate the effectiveness of $\Phi$, we compare the results obtained with and without the $\Phi$. Fig.~\ref{fig:quantitative_evaluation_of_the_sky_adjust_and_BrightnessPredNet} shows that the addition of $\Phi$ can maintain the quantitative evaluation performance. Fig.~\ref{fig:visual_evaluation_of_brightness_adjustment_and_BrightnessPredNet} demonstrates that the visual results of the image can be improved, which indicate the importance of $\Phi$.

\begin{figure}[!tb]
   \centering
   \footnotesize
   \includegraphics[width=2.0cm,height=1.4cm]{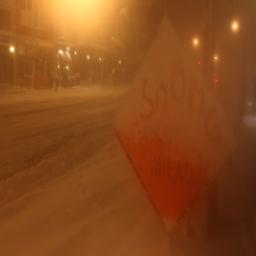}
   \includegraphics[width=2.0cm,height=1.4cm]{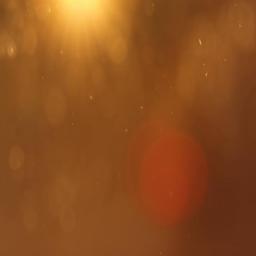}
   \includegraphics[width=2.0cm,height=1.4cm]{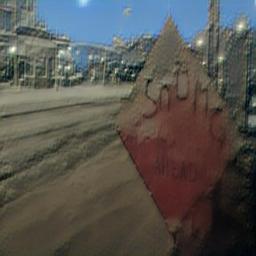}
   \includegraphics[width=2.0cm,height=1.4cm]{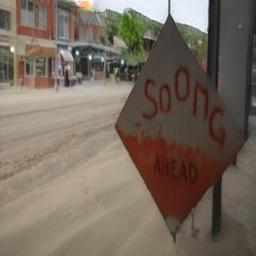}

   \includegraphics[width=2.0cm,height=1.4cm]{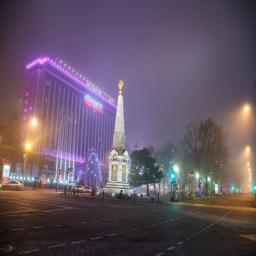}
   \includegraphics[width=2.0cm,height=1.4cm]{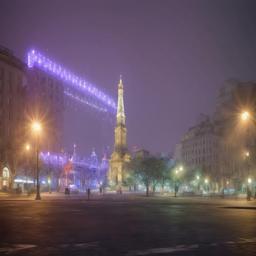}
   \includegraphics[width=2.0cm,height=1.4cm]{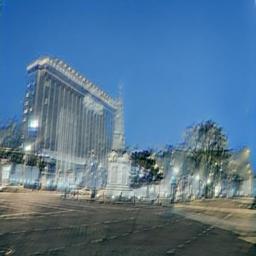}
   \includegraphics[width=2.0cm,height=1.4cm]{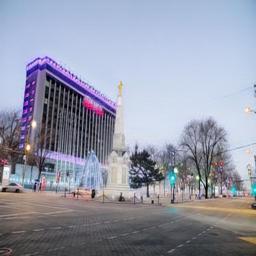}

   \leftline{\hspace{0.5cm} (a) Hazy \hspace{0.5cm} (b) w/o fine-tune \hspace{0.3cm} (c) w/o $\mathcal{L}_{adv}$ \hspace{0.6cm} (d) Ours}
   \caption{Visual evaluation of fine-tuning and the $\mathcal{L}_{adv}$.}
   \label{fig:visual_evaluation_of_fine_tuning_and_adversarial_loss}
 \end{figure}

\begin{figure}
   \footnotesize
   \centering
   \includegraphics[width=\linewidth]{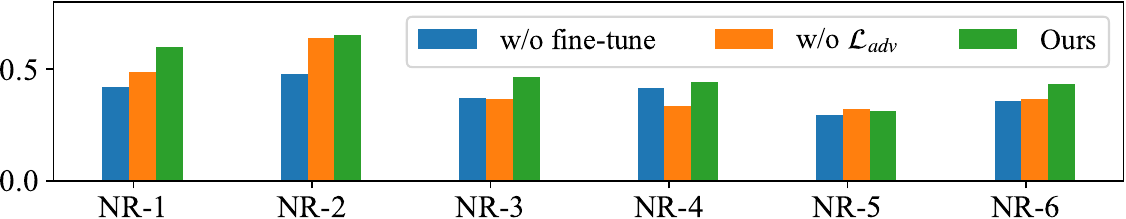}
   \caption{Quantitative evaluation of fine-tuning and $\mathcal{L}_{adv}$.}
   \label{fig:quantitative_evaluation_of_finetune_and_adversarial_loss}
 \end{figure}

\noindent \textbf{The pre-trained model and our fine-tuning.} 
The visual results obtained with and without fine-tuning on the diffusion model \cite{parmar2024one} are shown in Fig.~\ref{fig:visual_evaluation_of_fine_tuning_and_adversarial_loss}. The output image of the model shows an obvious hazy effect without fine-tuning. The results in Fig.~\ref{fig:quantitative_evaluation_of_finetune_and_adversarial_loss} show that the quantitative metrics obtained by the model without fine-tuning are lower. The main reason is that the original model does not consider the conditional injection process under our prompt and nighttime hazy images. The results demonstrate the necessity of our fine-tuning process.

\noindent \textbf{Analysis of the adversarial loss.}
During the training process, we use adversarial loss $\mathcal{L}_{adv}$ to optimize the generation results of the our model. To demonstrate the impact of the adversarial loss on image quality, the discriminator and the corresponding training process are removed. As shown in Fig.~\ref{fig:visual_evaluation_of_fine_tuning_and_adversarial_loss}, the structure and high-frequency information of the image are obviously lost without the adversarial loss. Meanwhile, the quantitative metrics shown in Fig.~\ref{fig:quantitative_evaluation_of_finetune_and_adversarial_loss} have also been slightly reduced. The results demonstrate the beneficial effect of adversarial loss on model performance.

\begin{figure}
   \footnotesize
   \centering
   \includegraphics[width=8.2cm,height=2cm]{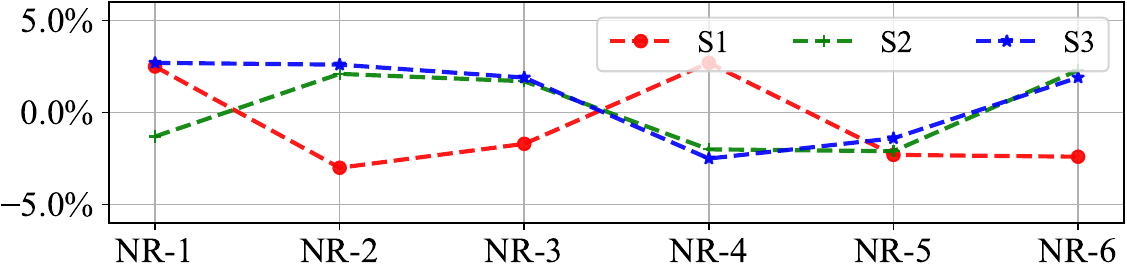}
   \caption{Quantitative evaluation of hyperparameters.}
   \label{fig:quantitative_evaluation_of_hyperparameters}
\end{figure}

\begin{figure}
   \centering
   \footnotesize
 
   \includegraphics[width=2.0cm,height=1.5cm]{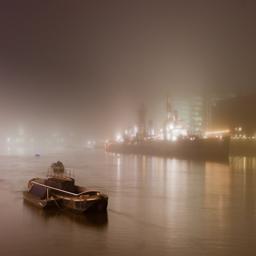}
   \includegraphics[width=2.0cm,height=1.5cm]{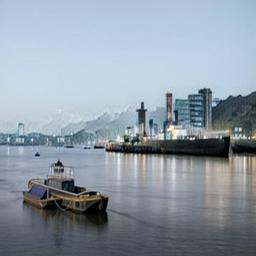}
   \includegraphics[width=2.0cm,height=1.5cm]{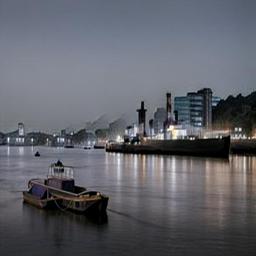}
   \includegraphics[width=2.0cm,height=1.5cm]{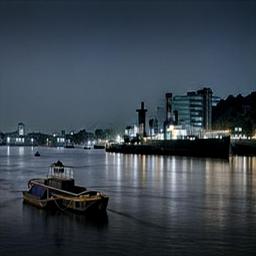}

   \includegraphics[width=2.02cm,height=1.5cm]{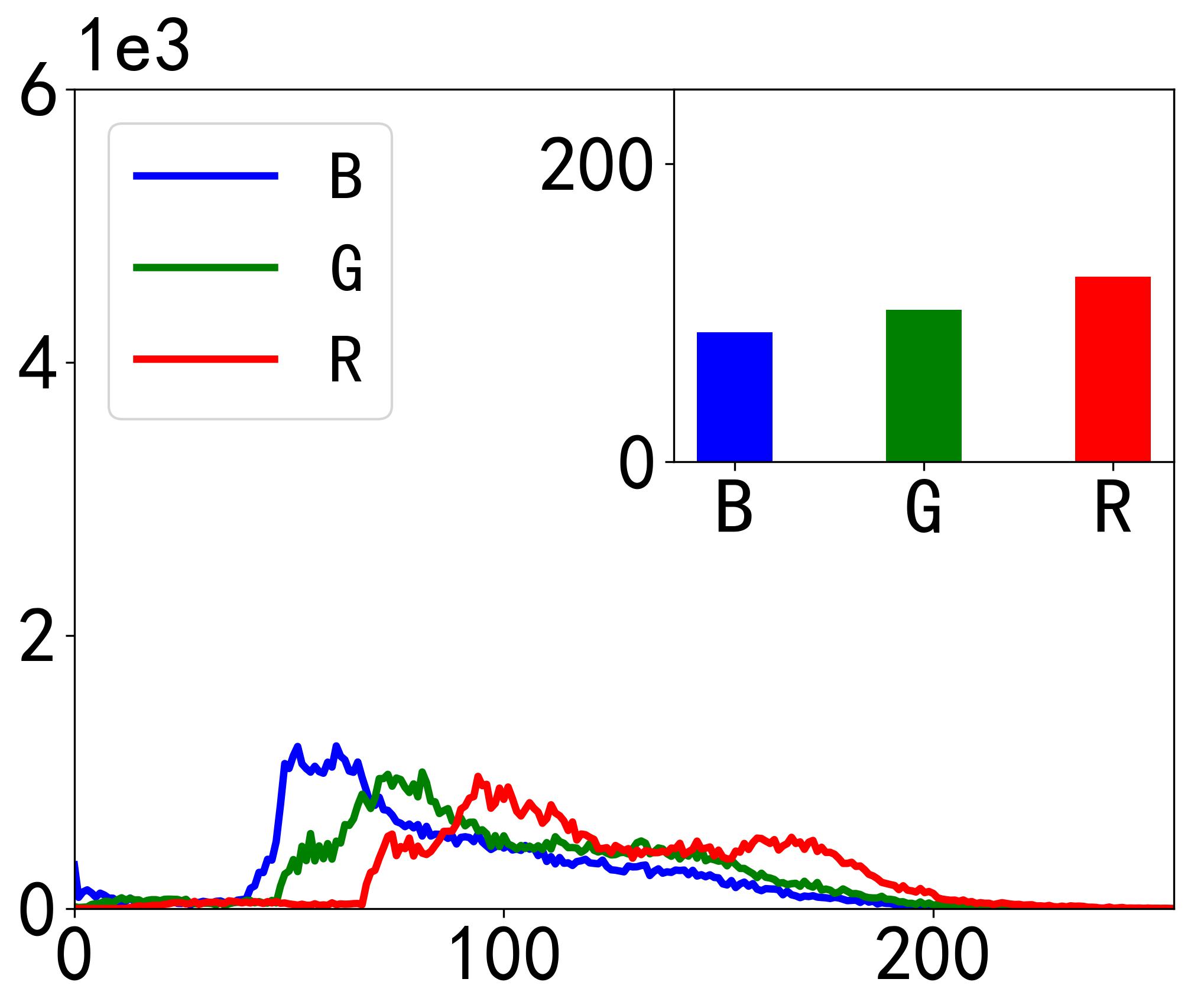}
   \includegraphics[width=2.0cm,height=1.5cm]{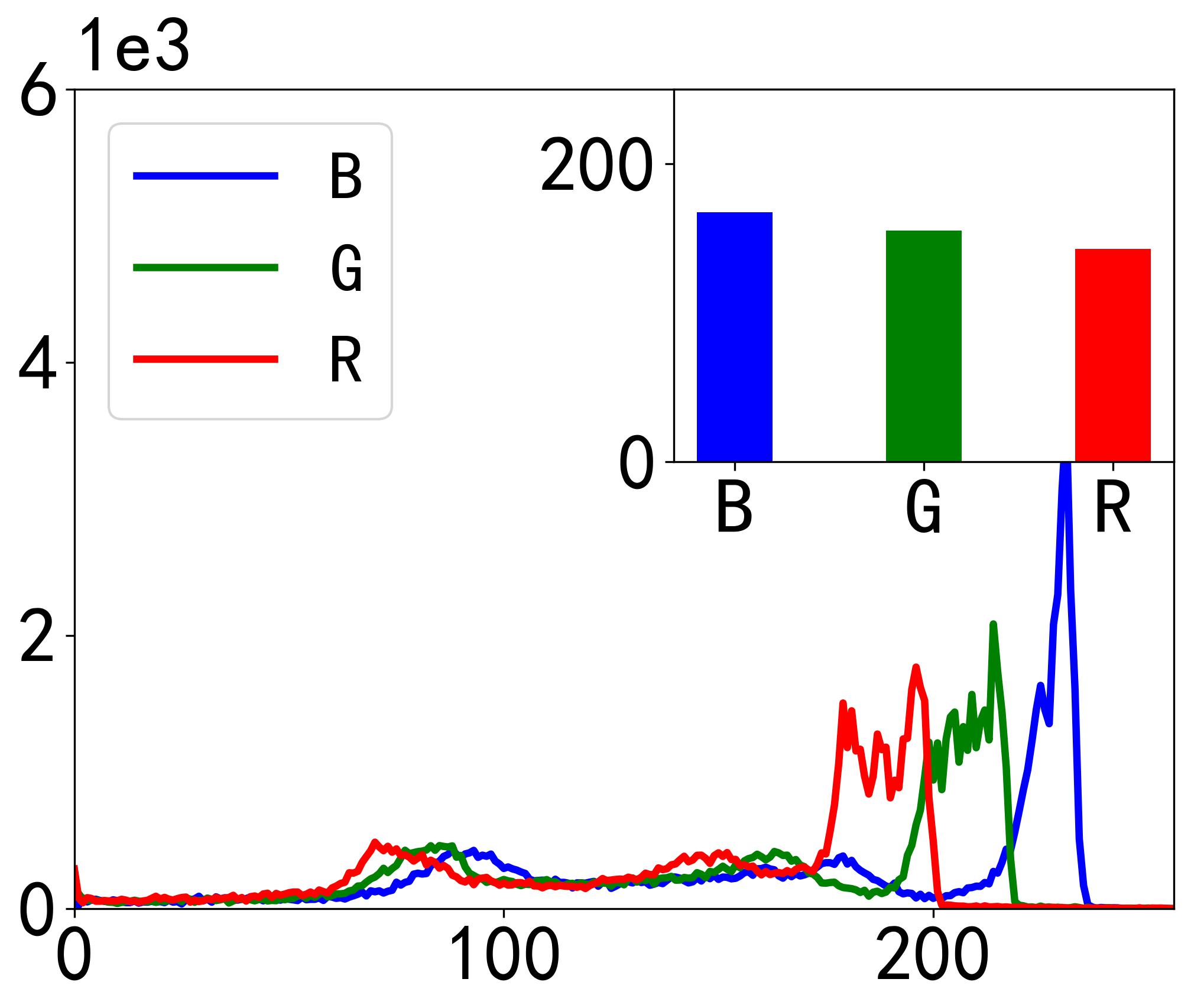}
   \includegraphics[width=2.0cm,height=1.5cm]{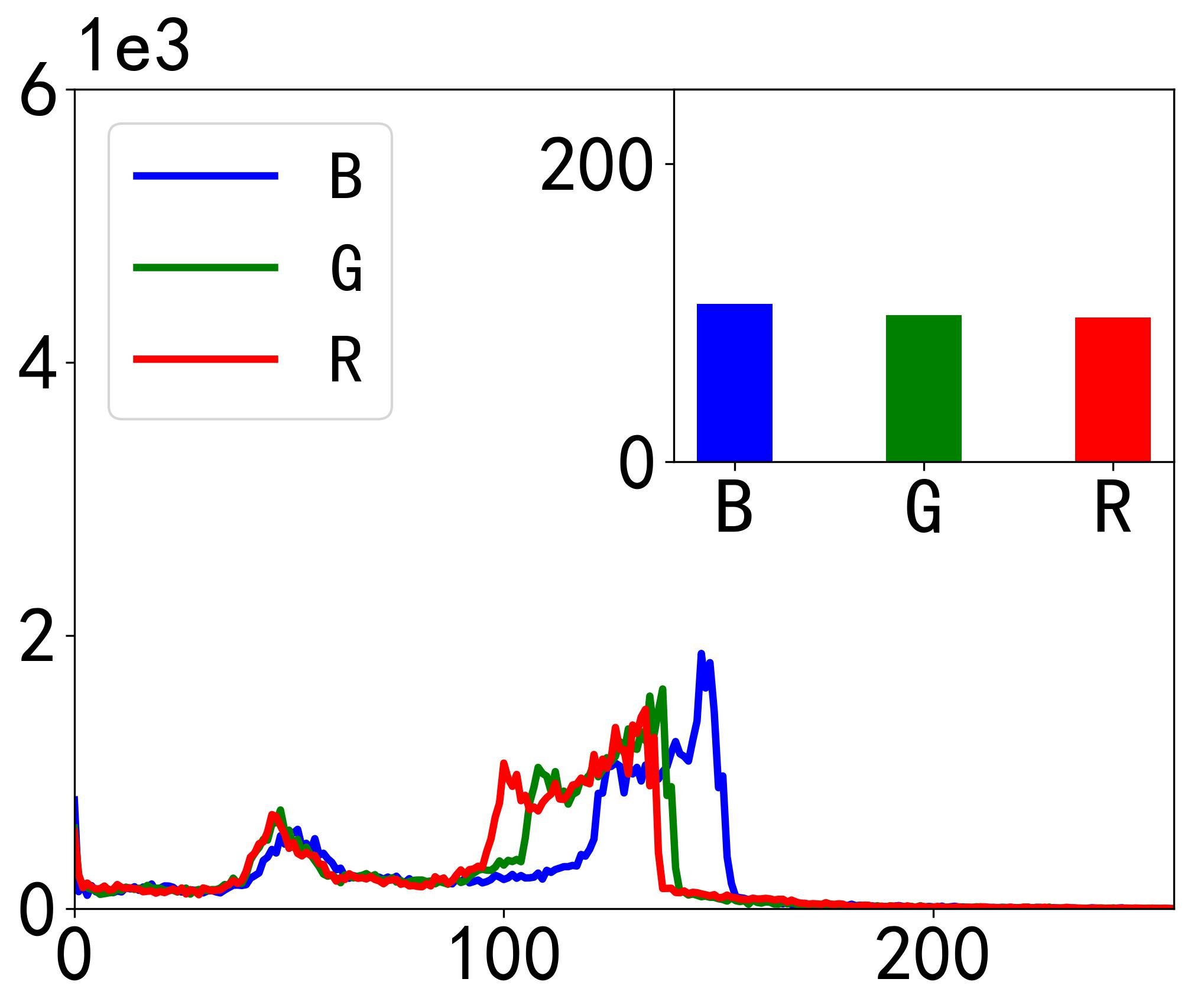}
   \includegraphics[width=2.0cm,height=1.5cm]{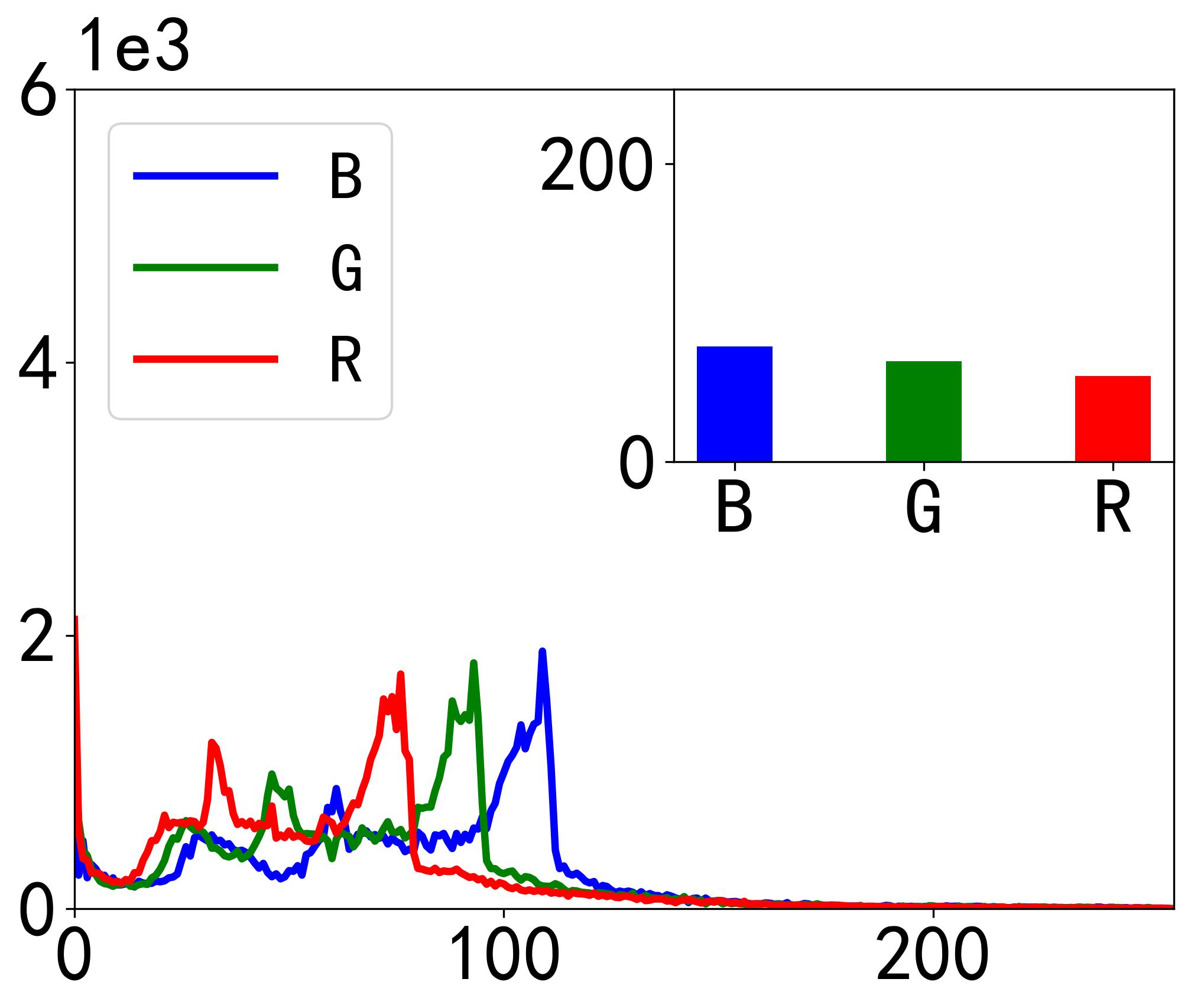}

   \leftline{\hspace{0.8cm} Hazy \hspace{0.6cm} a normal daytime  \hspace{0.1cm} a dusky evening \hspace{0.06cm} a dark nighttime}

   \caption{Results obtained with different prompts. By changing the prompts, our method can change the brightness to low-light situations. Each prompt in the sub-figures ends with ``image''.}
   \label{fig:comparison_of_differ_brightness_prompts}
 \end{figure}

\noindent \textbf{Analysis of hyperparameters.} Theoretically, the choices of hyperparameters are infinite. We empirically study the choices of hyperparameters quantitatively. While ensuring that the model's capabilities of image dehazing and night-to-day brightness mapping are not affected, the conducted settings include (S1) swapping the weight between $\lambda_{1}$ and $\lambda_{2}$, (S2) setting $\varrho$ to 0.975 (still able to segment the sky); (S3) setting VAE's LoRA rank to the same as UNet. The percentage changes in the different metrics given in Fig.~\ref{fig:quantitative_evaluation_of_hyperparameters} indicate that these hyperparameters have no significant impact on the performance.

\noindent \textbf{Analysis of Daytime and Nighttime Prompts.} During the training process, the prompt is fixed as ``a clear daytime image''. Since the representation space is only partially fine-tuned, the model retains its semantic understanding ability for the text input. We use different prompts to obtain the results for different lighting conditions. It is worth noting that low-lighting labels are not used during the training phase of our model. As shown in Fig.~\ref{fig:comparison_of_differ_brightness_prompts}, the content of the scene does not change when changing the image brightness, which proves that our training process can adequately maintain content consistency when the prompt is changed to different brightness description.

\section{Conclusion}
This paper introduces DiffND, a novel framework for nighttime dehazing and brightness mapping. We first propose a data synthesis pipeline that ensures consistent brightness patterns between synthetic and real-world scenes, facilitating the learning of night-to-day brightness mapping. Next, we present a diffusion-based dehazing model, guided by a brightness perception network, which injects brightness information into the decoding process to adapt the pre-trained diffusion model for high-quality lighting reconstruction. Our extensive validation demonstrates that both the dataset and the model are crucial for achieving superior performance. Quantitative and visual comparisons confirm the effectiveness of our approach in converting hazy nighttime images to clear, daytime-equivalent ones.


{\small
\bibliographystyle{ieee_fullname}
\bibliography{egbib}

\begin{thebibliography}{10}\itemsep=-1pt

\bibitem{agnolucci2024quality}
Lorenzo Agnolucci, Leonardo Galteri, and Marco Bertini.
\newblock Quality-aware image-text alignment for real-world image quality
  assessment.
\newblock {\em arXiv preprint arXiv:2403.11176}, 2024.

\bibitem{agnolucci2024arniqa}
Lorenzo Agnolucci, Leonardo Galteri, Marco Bertini, and Alberto Del~Bimbo.
\newblock Arniqa: Learning distortion manifold for image quality assessment.
\newblock In {\em IEEE WAVC}, pages 189--198, 2024.

\bibitem{banerjee2021nighttime}
Sriparna Banerjee and Sheli Sinha~Chaudhuri.
\newblock Nighttime image-dehazing: a review and quantitative benchmarking.
\newblock {\em Archives of Computational Methods in Engineering},
  28(4):2943--2975, 2021.

\bibitem{pyiqa}
Chaofeng Chen and Jiadi Mo.
\newblock {IQA-PyTorch}: Pytorch toolbox for image quality assessment.
\newblock [Online]. Available: \url{https://github.com/chaofengc/IQA-PyTorch},
  2022.

\bibitem{chen2024topiq}
Chaofeng Chen, Jiadi Mo, Jingwen Hou, Haoning Wu, Liang Liao, Wenxiu Sun, Qiong
  Yan, and Weisi Lin.
\newblock Topiq: A top-down approach from semantics to distortions for image
  quality assessment.
\newblock {\em IEEE TIP}, 2024.

\bibitem{chen2024ni}
Hui Chen, Nannan Li, and Rong Chen.
\newblock Ni-dehazenet: representation learning via bilevel optimized
  architecture search for nighttime dehazing.
\newblock {\em The Visual Computer}, 40(9):6155--6170, 2024.

\bibitem{chung2024style}
Jiwoo Chung, Sangeek Hyun, and Jae-Pil Heo.
\newblock Style injection in diffusion: A training-free approach for adapting
  large-scale diffusion models for style transfer.
\newblock In {\em IEEE CVPR}, pages 8795--8805, 2024.

\bibitem{cong2024semi}
Xiaofeng Cong, Jie Gui, Jing Zhang, Junming Hou, and Hao Shen.
\newblock A semi-supervised nighttime dehazing baseline with spatial-frequency
  aware and realistic brightness constraint.
\newblock In {\em IEEE CVPR}, pages 2631--2640, 2024.

\bibitem{cui2023focal}
Yuning Cui, Wenqi Ren, Xiaochun Cao, and Alois Knoll.
\newblock Focal network for image restoration.
\newblock In {\em IEEE ICCV}, pages 13001--13011, 2023.

\bibitem{cui2024omni}
Yuning Cui, Wenqi Ren, and Alois Knoll.
\newblock Omni-kernel network for image restoration.
\newblock In {\em AAAI}, volume~38, pages 1426--1434, 2024.

\bibitem{dai2022flare7k}
Yuekun Dai, Chongyi Li, Shangchen Zhou, Ruicheng Feng, and Chen~Change Loy.
\newblock Flare7k: A phenomenological nighttime flare removal dataset.
\newblock {\em NeurIPS}, 35:3926--3937, 2022.

\bibitem{golestaneh2021no}
S~Alireza Golestaneh, Saba Dadsetan, and Kris~M Kitani.
\newblock No-reference image quality assessment via transformers, relative
  ranking, and self-consistency.
\newblock In {\em IEEE WACV}, pages 3209--3218, 2022.

\bibitem{hu2021lora}
Edward~J Hu, Yelong Shen, Phillip Wallis, Zeyuan Allen-Zhu, Yuanzhi Li, Shean
  Wang, Lu Wang, and Weizhu Chen.
\newblock Lora: Low-rank adaptation of large language models.
\newblock {\em arXiv preprint arXiv:2106.09685}, 2021.

\bibitem{jin2025raindrop}
Yeying Jin, Xin Li, Jiadong Wang, Yan Zhang, and Malu Zhang.
\newblock Raindrop clarity: A dual-focused dataset for day and night raindrop
  removal.
\newblock In {\em ECCV}, pages 1--17, 2024.

\bibitem{jin2023enhancing}
Yeying Jin, Beibei Lin, Wending Yan, Wei Ye, Yuan Yuan, and Robby~T Tan.
\newblock Enhancing visibility in nighttime haze images using guided apsf and
  gradient adaptive convolution.
\newblock In {\em ACM MM}, 2023.

\bibitem{jin2022structure}
Yeying Jin, Wending Yan, Wenhan Yang, and Robby~T Tan.
\newblock Structure representation network and uncertainty feedback learning
  for dense non-uniform fog removal.
\newblock In {\em ACCV}, pages 155--172. Springer, 2022.

\bibitem{jin2022unsupervised}
Yeying Jin, Wenhan Yang, and Robby~T Tan.
\newblock Unsupervised night image enhancement: When layer decomposition meets
  light-effects suppression.
\newblock In {\em ECCV}, pages 404--421. Springer, 2022.

\bibitem{ke2024repurposing}
Bingxin Ke, Anton Obukhov, Shengyu Huang, Nando Metzger, Rodrigo~Caye Daudt,
  and Konrad Schindler.
\newblock Repurposing diffusion-based image generators for monocular depth
  estimation.
\newblock In {\em IEEE CVPR}, pages 9492--9502, 2024.

\bibitem{ke2021musiq}
Junjie Ke, Qifei Wang, Yilin Wang, Peyman Milanfar, and Feng Yang.
\newblock Musiq: Multi-scale image quality transformer.
\newblock In {\em IEEE ICCV}, pages 5148--5157, 2021.

\bibitem{kennerley20232pcnet}
Mikhail Kennerley, Jian-Gang Wang, Bharadwaj Veeravalli, and Robby~T Tan.
\newblock 2pcnet: Two-phase consistency training for day-to-night unsupervised
  domain adaptive object detection.
\newblock In {\em IEEE CVPR}, pages 11484--11493, 2023.

\bibitem{kuanar2022multi}
Shiba Kuanar, Dwarikanath Mahapatra, Monalisa Bilas, and KR Rao.
\newblock Multi-path dilated convolution network for haze and glow removal in
  nighttime images.
\newblock {\em The Visual Computer}, pages 1--14, 2022.

\bibitem{kumari2022ensembling}
Nupur Kumari, Richard Zhang, Eli Shechtman, and Jun-Yan Zhu.
\newblock Ensembling off-the-shelf models for gan training.
\newblock In {\em IEEE CVPR}, pages 10651--10662, 2022.

\bibitem{li2018benchmarking}
Boyi Li, Wenqi Ren, Dengpan Fu, Dacheng Tao, Dan Feng, Wenjun Zeng, and
  Zhangyang Wang.
\newblock Benchmarking single-image dehazing and beyond.
\newblock {\em IEEE TIP}, 28(1):492--505, 2018.

\bibitem{li2024learning}
Tianqi Li, Guansong Pang, Xiao Bai, Wenjun Miao, and Jin Zheng.
\newblock Learning transferable negative prompts for out-of-distribution
  detection.
\newblock In {\em IEEE CVPR}, pages 17584--17594, 2024.

\bibitem{li2015nighttime}
Yu Li, Robby~T Tan, and Michael~S Brown.
\newblock Nighttime haze removal with glow and multiple light colors.
\newblock In {\em IEEE ICCV}, pages 226--234, 2015.

\bibitem{liao2018hdp}
Yinghong Liao, Zhuo Su, Xiangguo Liang, and Bin Qiu.
\newblock Hdp-net: Haze density prediction network for nighttime dehazing.
\newblock In {\em Pacific Rim Conference on Multimedia}, pages 469--480, 2018.

\bibitem{lin2024nighthaze}
Beibei Lin, Yeying Jin, Wending Yan, Wei Ye, Yuan Yuan, and Robby~T Tan.
\newblock Nighthaze: Nighttime image dehazing via self-prior learning.
\newblock In {\em AAAI}, 2025.

\bibitem{lin2024nightrain}
Beibei Lin, Yeying Jin, Wending Yan, Wei Ye, Yuan Yuan, Shunli Zhang, and
  Robby~T Tan.
\newblock Nightrain: Nighttime video deraining via adaptive-rain-removal and
  adaptive-correction.
\newblock In {\em AAAI}, volume~38, pages 3378--3385, 2024.

\bibitem{liu2021single}
Yun Liu, Anzhi Wang, Hao Zhou, and Pengfei Jia.
\newblock Single nighttime image dehazing based on image decomposition.
\newblock {\em Signal Processing}, 183:107986, 2021.

\bibitem{liu2023nighthazeformer}
Yun Liu, Zhongsheng Yan, Sixiang Chen, Tian Ye, Wenqi Ren, and Erkang Chen.
\newblock Nighthazeformer: Single nighttime haze removal using prior query
  transformer.
\newblock In {\em ACM MM}, 2023.

\bibitem{liu2022multi}
Yun Liu, Zhongsheng Yan, Jinge Tan, and Yuche Li.
\newblock Multi-purpose oriented single nighttime image haze removal based on
  unified variational retinex model.
\newblock {\em IEEE TCSVT}, 33(4):1643--1657, 2022.

\bibitem{liu2022nighttime}
Yun Liu, Zhongsheng Yan, Aimin Wu, Tian Ye, and Yuche Li.
\newblock Nighttime image dehazing based on variational decomposition model.
\newblock In {\em IEEE CVPR Workshops}, pages 640--649, 2022.

\bibitem{liu2022single}
Yun Liu, Zhongsheng Yan, Tian Ye, Aimin Wu, and Yuche Li.
\newblock Single nighttime image dehazing based on unified variational
  decomposition model and multi-scale contrast enhancement.
\newblock {\em Engineering Applications of Artificial Intelligence},
  116:105373, 2022.

\bibitem{parmar2023zero}
Gaurav Parmar, Krishna Kumar~Singh, Richard Zhang, Yijun Li, Jingwan Lu, and
  Jun-Yan Zhu.
\newblock Zero-shot image-to-image translation.
\newblock In {\em ACM SIGGRAPH}, pages 1--11, 2023.

\bibitem{parmar2024one}
Gaurav Parmar, Taesung Park, Srinivasa Narasimhan, and Jun-Yan Zhu.
\newblock One-step image translation with text-to-image models.
\newblock {\em arXiv preprint arXiv:2403.12036}, 2024.

\bibitem{rombach2022high}
Robin Rombach, Andreas Blattmann, Dominik Lorenz, Patrick Esser, and Bjorn
  Ommer.
\newblock High-resolution image synthesis with latent diffusion models.
\newblock In {\em IEEE CVPR}, pages 10684--10695, 2022.

\bibitem{sauer2024adversarial}
Axel Sauer, Dominik Lorenz, Andreas Blattmann, and Robin Rombach.
\newblock Adversarial diffusion distillation.
\newblock In {\em ECCV}, pages 87--103, 2024.

\bibitem{sharma2020single}
Aashish Sharma, Robby~T Tan, and Loong-Fah Cheong.
\newblock Single-image camera response function using prediction consistency
  and gradual refinement.
\newblock In {\em ACCV}, 2020.

\bibitem{si2024freeu}
Chenyang Si, Ziqi Huang, Yuming Jiang, and Ziwei Liu.
\newblock Freeu: Free lunch in diffusion u-net.
\newblock In {\em IEEE CVPR}, pages 4733--4743, 2024.

\bibitem{tian2024argue}
Xinyu Tian, Shu Zou, Zhaoyuan Yang, and Jing Zhang.
\newblock Argue: Attribute-guided prompt tuning for vision-language models.
\newblock In {\em IEEE CVPR}, pages 28578--28587, 2024.

\bibitem{wang2022exploring}
Jianyi Wang, Kelvin~CK Chan, and Chen~Change Loy.
\newblock Exploring clip for assessing the look and feel of images.
\newblock In {\em AAAI}, 2023.

\bibitem{wang2022variational}
Wenhui Wang, Anna Wang, and Chen Liu.
\newblock Variational single nighttime image haze removal with a gray haze-line
  prior.
\newblock {\em IEEE TIP}, 31:1349--1363, 2022.

\bibitem{wei2024robust_2}
Haoran Wei, Qingbo Wu, Chenhao Wu, Shuai Chen, Lei Wang, King~Ngi Ngan, Fanman
  Meng, and Hongliang Li.
\newblock Robust real-world image dehazing via knowledge guided conditional
  diffusion model finetuning.
\newblock In {\em IEEE Workshop on MMSP}, pages 1--6. IEEE, 2024.

\bibitem{wei2024robust}
Haoran Wei, Qingbo Wu, Chenhao Wu, King~Ngi Ngan, Hongliang Li, Fanman Meng,
  and Heqian Qiu.
\newblock Robust unpaired image dehazing via adversarial deformation
  constraint.
\newblock {\em IEEE TCSVT}, 2024.

\bibitem{wu2023q}
Haoning Wu, Zicheng Zhang, Weixia Zhang, Chaofeng Chen, Liang Liao, Chunyi Li,
  Yixuan Gao, Annan Wang, Erli Zhang, Wenxiu Sun, et~al.
\newblock Q-align: Teaching lmms for visual scoring via discrete text-defined
  levels.
\newblock {\em arXiv preprint arXiv:2312.17090}, 2023.

\bibitem{wu2024comprehensive}
Tianhe Wu, Kede Ma, Jie Liang, Yujiu Yang, and Lei Zhang.
\newblock A comprehensive study of multimodal large language models for image
  quality assessment.
\newblock In {\em ECCV}, pages 143--160. Springer, 2024.

\bibitem{xu2024prompt}
Xingqian Xu, Jiayi Guo, Zhangyang Wang, Gao Huang, Irfan Essa, and Humphrey
  Shi.
\newblock Prompt-free diffusion: Taking text out of text-to-image diffusion
  models.
\newblock In {\em IEEE CVPR}, pages 8682--8692, 2024.

\bibitem{yan2024diffusion}
Jing~Nathan Yan, Jiatao Gu, and Alexander~M Rush.
\newblock Diffusion models without attention.
\newblock In {\em IEEE CVPR}, pages 8239--8249, 2024.

\bibitem{yan2020nighttime}
Wending Yan, Robby~T Tan, and Dengxin Dai.
\newblock Nighttime defogging using high-low frequency decomposition and
  grayscale-color networks.
\newblock In {\em ECCV}, pages 473--488, 2020.

\bibitem{yang2022variation}
Chih-Hsiang Yang, Yi-Hsien Lin, and Yi-Chang Lu.
\newblock A variation-based nighttime image dehazing flow with a physically
  valid illumination estimator and a luminance-guided coloring model.
\newblock {\em IEEE Access}, 10:50153--50166, 2022.

\bibitem{depth_anything_v2}
Lihe Yang, Bingyi Kang, Zilong Huang, Zhen Zhao, Xiaogang Xu, Jiashi Feng, and
  Hengshuang Zhao.
\newblock Depth anything v2.
\newblock {\em arXiv:2406.09414}, 2024.

\bibitem{yang2018superpixel}
Minmin Yang, Jianchang Liu, and Zhengguo Li.
\newblock Superpixel-based single nighttime image haze removal.
\newblock {\em IEEE TMM}, 20(11):3008--3018, 2018.

\bibitem{yang2022maniqa}
Sidi Yang, Tianhe Wu, Shuwei Shi, Shanshan Lao, Yuan Gong, Mingdeng Cao, Jiahao
  Wang, and Yujiu Yang.
\newblock Maniqa: Multi-dimension attention network for no-reference image
  quality assessment.
\newblock In {\em IEEE CVPR}, pages 1191--1200, 2022.

\bibitem{yin2024one}
Tianwei Yin, Micha{\"e}l Gharbi, Richard Zhang, Eli Shechtman, Fredo Durand,
  William~T Freeman, and Taesung Park.
\newblock One-step diffusion with distribution matching distillation.
\newblock In {\em IEEE CVPR}, pages 6613--6623, 2024.

\bibitem{yu2019nighttime}
Teng Yu, Kang Song, Pu Miao, Guowei Yang, Huan Yang, and Chenglizhao Chen.
\newblock Nighttime single image dehazing via pixel-wise alpha blending.
\newblock {\em IEEE Access}, 7:114619--114630, 2019.

\bibitem{zhang2017fast}
Jing Zhang, Yang Cao, Shuai Fang, Yu Kang, and Chang Wen~Chen.
\newblock Fast haze removal for nighttime image using maximum reflectance
  prior.
\newblock In {\em IEEE CVPR}, pages 7418--7426, 2017.

\bibitem{zhang2014nighttime}
Jing Zhang, Yang Cao, and Zengfu Wang.
\newblock Nighttime haze removal based on a new imaging model.
\newblock In {\em IEEE ICIP}, pages 4557--4561, 2014.

\bibitem{zhang2020nighttime}
Jing Zhang, Yang Cao, Zheng-Jun Zha, and Dacheng Tao.
\newblock Nighttime dehazing with a synthetic benchmark.
\newblock In {\em ACM MM}, pages 2355--2363, 2020.

\bibitem{zhang2020blind}
Weixia Zhang, Kede Ma, Jia Yan, Dexiang Deng, and Zhou Wang.
\newblock Blind image quality assessment using a deep bilinear convolutional
  neural network.
\newblock {\em IEEE TCSVT}, 30(1):36--47, 2020.

\bibitem{zhang2023blind}
Weixia Zhang, Guangtao Zhai, Ying Wei, Xiaokang Yang, and Kede Ma.
\newblock Blind image quality assessment via vision-language correspondence: A
  multitask learning perspective.
\newblock In {\em IEEE CVPR}, pages 14071--14081, 2023.

\bibitem{zhou2017places}
Bolei Zhou, Agata Lapedriza, Aditya Khosla, Aude Oliva, and Antonio Torralba.
\newblock Places: A 10 million image database for scene recognition.
\newblock {\em IEEE TPAMI}, 40(6):1452--1464, 2017.

\bibitem{zhou2023fourmer}
Man Zhou, Jie Huang, Chun-Le Guo, and Chongyi Li.
\newblock Fourmer: an efficient global modeling paradigm for image restoration.
\newblock In {\em ICML}, pages 42589--42601, 2023.

\bibitem{zhou2022edge}
Wujie Zhou, Shaohua Dong, Caie Xu, and Yaguan Qian.
\newblock Edge-aware guidance fusion network for rgb--thermal scene parsing.
\newblock In {\em AAAI}, volume~36, pages 3571--3579, 2022.

\end{thebibliography}
}

\end{document}